%% file: bare_jrnl_new_sample4.tex
\documentclass[lettersize,journal]{IEEEtran}
\usepackage{amsmath,amsfonts}
\usepackage{algorithm}
\usepackage{array}
\usepackage{marvosym}
\usepackage[caption=false,font=normalsize,labelfont=sf,textfont=sf]{subfig}
\usepackage{textcomp}
\usepackage{stfloats}
\usepackage{url}
\usepackage{ulem}
\usepackage{verbatim}
\usepackage{makecell}
\usepackage{graphicx}
\usepackage{cite}
\usepackage{algpseudocode}  
\usepackage{multirow}
\usepackage{adjustbox}
\usepackage{xcolor}
\usepackage{booktabs}
\usepackage{colortbl}
\definecolor{cvprblue}{rgb}{0.21,0.49,0.74}
\definecolor{tabhighlight}{HTML}{e5e5e5}
\definecolor{grey}{RGB}{128,138,135}
\definecolor{oorange}{RGB}{215,122,71}
\definecolor{yyellow}{RGB}{230,185,79}
\definecolor{ppurple}{RGB}{122,30,97}
\definecolor{ggreen}{RGB}{112,173,71}
\definecolor{battleshipgrey}{rgb}{0.3, 0.3, 0.3}
\definecolor{brilliantrose}{rgb}{1.0, 0.33, 0.64}
\definecolor{americanrose}{rgb}{1.0, 0.01, 0.24}
\definecolor{jweigreen}{rgb}{0,0.45,0.24}
\definecolor{bluegray}{rgb}{0.1, 0.1, 0.4}
\definecolor{ao(english)}{rgb}{0.0, 0.5, 0.0}
\definecolor{blanchedalmond}{rgb}{1.0, 0.92, 0.8}
\definecolor{atomictangerine}{rgb}{1.0, 0.6, 0.4}
\definecolor{chocolate(web)}{rgb}{0.82, 0.41, 0.12}
\definecolor{bananayellow}{rgb}{1.0, 0.88, 0.21}
\definecolor{goldenbrown}{rgb}{0.6, 0.4, 0.08}
\definecolor{aliceblue}{rgb}{0.94, 0.97, 1.0}
\definecolor{beige}{rgb}{0.96, 0.96, 0.86}
\definecolor{babyblue}{rgb}{0.54, 0.81, 0.94}
\definecolor{camel}{rgb}{0.76, 0.6, 0.42}
\definecolor{cinnamon}{rgb}{0.82, 0.41, 0.12}
\definecolor{redlinkcolor}{rgb}{0.79607843, 0.25098039, 0.25882353}
\definecolor{bluecitecolor}{rgb}{0,0.36,0.69}
\usepackage[pagebackref,breaklinks,colorlinks,allcolors=cvprblue]{hyperref}

\usepackage{tabularx}

\begin{document}

\title{WAT: Online Video Understanding Needs Watching Before Thinking}

\author{Zifan~Han,
       Hongbo~Sun,
       Jinglin~Xu,
       Canhui~Tang,
       Yulong~Lei,
       Xuchong~Zhang,
       Hongbin~Sun,
       Zhongjiang He,
       Hao~Sun

       \thanks{Zifan Han, Canhui Tang, Xuchong Zhang, and Hongbin Sun are with the National Key Laboratory of Human-Machine Hybrid Augmented Intelligence, Institute of Artificial Intelligence and Robotics, Xi’an Jiaotong University, Xi'an 710049, P. R. China(e-mail: hzf724298913@stu.xjtu.edu.cn; 2193512014@stu.xjtu.edu.cn; zhangxc0329@xjtu.edu.cn; hsun@mail.xjtu.edu.cn).}
       \thanks{Hongbo Sun, Zhongjiang He, and Hao Sun are with the Institute of Artificial Intelligence (TeleAI), China Telecom,  31 Jinrong Ave., Xicheng District, Beijing 100033, P. R. China(e-mail: sunhb3@chinatelecom.cn; hezj@chinatelecom.cn; sunh10@chinatelecom.cn).}
       \thanks{Jinglin Xu, and Yulong Lei are with the University of Science and Technology Beijing, Beijing 100083, P. R. China(e-mail: xujinglinlove@gmail.com; m202420916@xs.ustb.edu.cn).}
}




\maketitle

\begin{abstract}
Multimodal Large Language Models (MLLMs) have demonstrated strong capabilities in image understanding. Building on this success, significant efforts have recently been devoted to extending these models to video understanding, giving rise to the field of Video LLMs. However, vanilla Video LLMs' performance in online streaming scenarios remains severely limited by the need to preserve rich historical context under strict memory constraints while maintaining temporal coherence over long, continuous streams. To address these issues, we propose WAT (Watching Before Thinking), a new two-stage Video LLM framework designed for online video reasoning. WAT decouples processing into a query-independent watching stage and a query-triggered thinking stage. In the Watching stage, we establish a hierarchical memory system. It consists of a Short-Term Memory (STM) that serves as a high‑fidelity buffer for continuously caching recent frames and a fixed‑capacity Long‑Term Memory (LTM) that maintains a semantically diverse summary of historical content. The LTM is governed by a redundancy‑aware eviction policy, which selectively retains distinctive visual information across hundreds of frames. In the thinking stage, a context-aware retrieval mechanism first fuses the query with the current STM context and then retrieves the most relevant historical frames from the LTM; the combined evidence, further aligned via Retrieval Alignment Contrastive Learning (RACL) during training, enables coherent cross-temporal reasoning in the MLLM. In addition, to address data scarcity in genuine online video tasks, we introduce WAT-85K, a curated training dataset that integrates offline sources with high-quality streaming-style annotations that emphasize real-time perception, backward tracing, and proactive forecasting. Experiments show that WAT achieves state-of-the-art performance on online video benchmarks, including 77.7\% accuracy on StreamingBench and 55.2\% accuracy on OVO-Bench, significantly outperforming all existing open-source online Video LLMs. Moreover, WAT matches or surpasses strong offline MLLMs while operating at real-time frame rates, demonstrating that WAT is a robust, efficient, and generalizable solution for online video understanding.
\end{abstract}


\section{Introduction}
\label{sec:intro}

Multimodal Large Language Models (MLLMs)~\cite{an2026aiflow, li2024multiagent, liu2025trainingfree, shen2025lvgvos, yu2024multi, fu2025boosting, xie2025caption, yellinek20253vl, ataallah2024minigpt4, bai2025qwen25vltechnicalreport, li2025videochat, li2024temporal, liu2024llavanext, ren2024timechat, zhang2025videollama, zhang2024internlm, zhang2024video, xu2024slowfast} have made significant progress in video understanding, demonstrating strong capabilities in semantic comprehension and temporal reasoning across diverse video content. As a natural extension, applying these models to online streaming scenarios has attracted great attention, where videos are processed incrementally as frames arrive in real time. Growing demands from applications such as autonomous driving~\cite{shao2024lmdrive}, live video analysis~\cite{chen2025livecc}, and robotic interaction~\cite{zhu2024spa} necessitate a shift from conventional offline video understanding to the more challenging paradigm.

Existing methods~\cite{zhang2024internlm, chen2024videollm, di2025streaming, qian2024streaming, xu2025qwen2, zhang2024flash} for online video understanding must employ feature compression to handle the continuous influx of video frames under finite computational resources. The core challenge~\cite{chen2024videollm, di2025streaming, qian2024streaming, zhang2024flash, yao2025timechat} is determining the optimal compression strategy that preserves both fine-grained temporal details and long-term semantic coherence while minimizing computational overhead. Most existing approaches can be categorized into two types:

\begin{figure}[t]
\includegraphics[width=0.50\textwidth]{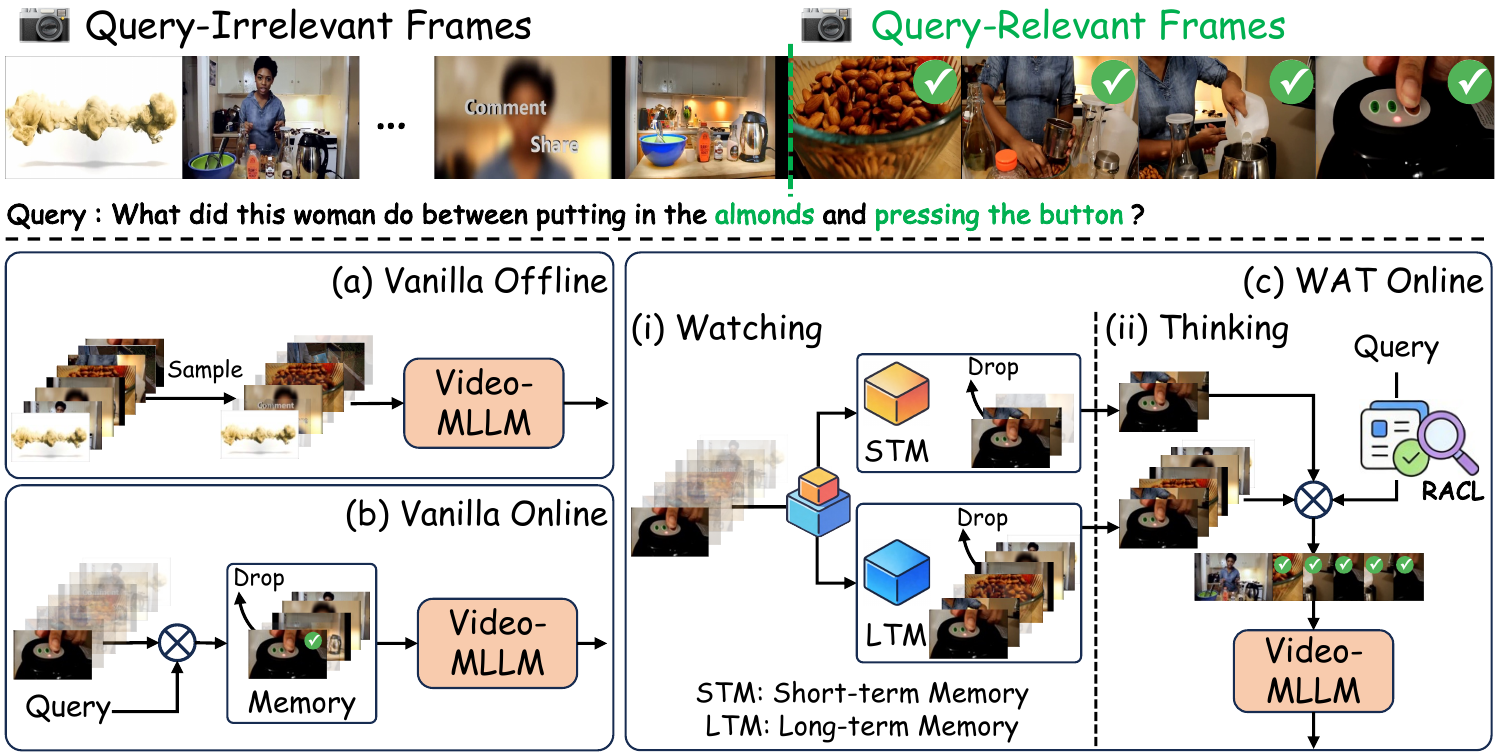}
  \caption{Illustrations of different Video LLMs: (a) Vanilla offline processing, which encodes the uniformly sampled frames at once; (b) Vanilla online processing, which encodes frames sequentially into the space-limited memory; while (c) Our WAT introduces a \textit{watching-before-thinking} pipeline to enhance online comprehension through hierarchical memory and explicit query-based retrieval reasoning.} 
   \label{pipe}\
\end{figure}

\begin{itemize}
    \item \textbf{Sampling-time compression} aggressively prunes incoming frames, reducing computational cost but severely limiting the model's capacity for fine-grained spatiotemporal reasoning. As a result, these methods often degrade into coarse scene summarization, failing to capture transient or fine-grained activities.
    
    \item  \textbf{Storage-time compression} retains more frames but merges or discards adjacent ones post-hoc based on feature similarity. While more memory-efficient in practice, this strategy risks missing critical foreground activities when they are overshadowed by highly dominant background content. Moreover, excessive local merging can severely disrupt temporal coherence, thereby impairing the fidelity of long-term reasoning
\end{itemize}

These limitations highlight a fundamental tension in online video understanding: how can we preserve both fine-grained temporal contextual information and long-term semantic diversity without exceeding computational budget and memory storage limitations? This raises a crucial question: Can we develop an efficient memory architecture that enables sustained reasoning over continuous video streams while maintaining both temporal coherence and semantic diversity? However, there are two fundamental challenges in designing such an architecture: 
(1) Memory Efficiency: how to maintain a compact yet comprehensive feature cache spanning hundreds of frames without redundant storage. (2) Temporal Coherence: how to ensure that the preserved features maintain accurate temporal relationships, which are critical for reasoning about actions and events.

To this end, we propose \textbf{WAT (Watching Before Thinking)}, new two-stage online Video LLM framework based on a simple yet powerful principle: before reasoning, the model must first "watch" the video effectively. 
Specifically, WAT comprises two main stages: the watching stage and the thinking stage. The watching stage addresses the memory-efficiency challenge through a two-level hierarchical design: a Short-Term Memory (STM) that preserves high-fidelity recent context via a sliding window, and a Long-Term Memory (LTM) that maintains a compact, diverse set of historically salient frames. The redundancy-aware eviction policy in LTM ensures memory efficiency by selectively replacing over-represented features based on cosine similarity, while protecting recent content from premature deletion.
For the temporal coherence challenge, WAT employs distinct update strategies for different memory components. The STM continuously buffers recent frames to preserve fine-grained immediate temporal context. In contrast, the LTM is updated to reflect information from the entire video. When the LTM reaches its storage limit, it admits new frames while evicting the frame most similar to those already in memory. This LTM update mechanism ensures the long-term store maintains a rich and varied representation of the video's evolution. 

The Thinking stage is triggered by a textual query. At this stage, WAT employs a context-aware retrieval mechanism: it first fuses the query with the current Short-Term Memory (\(\mathcal{M}_S\)) to form a conditioned query vector, which is then used to retrieve the most relevant historical features from the Long-Term Memory (\(\mathcal{M}_L\)). The integrated visual evidence—combining the immediate context from \(\mathcal{M}_S\) with the retrieved content—is then fed into the Multimodal Large Language Model (MLLM) for coherent, cross-temporal reasoning.
During training, the retrieval process is further refined by the auxiliary Retrieval Alignment Contrastive Learning (RACL) objective. This self-supervised loss aligns query-guided retrieval with the video's underlying semantics, thereby enhancing the relevance and accuracy of selected long-term memories.
Extensive experiments validate the effectiveness of WAT, achieving state-of-the-art performance on online video benchmarks, with 77.7\% on StreamingBench and 55.2\% on OVO-Bench, and matching or surpassing leading offline MLLMs in offline evaluations.

To enable effective training of such online Video LLMs, we introduce WAT-85K, a high-quality dataset specifically designed and augmented for streaming-style video question answering. Unlike existing video-language datasets, which are largely optimized for offline reasoning over short or pre-segmented clips, In particlar, WAT-85K emphasizes real-time perception and proactive forecasting, two core online reasoning capabilities that are underrepresented in prior datasets.

Our contributions can be summarized as follows:

\begin{itemize}
    \item We propose a new two-stage online Video LLM framework, WAT. This framework comprises two parts: watching and thinking. The watching part provides comprehensive visual perception, while the thinking part integrates with the problem to achieve comprehension. 

    \item We construct a hierarchical memory with Short-Term Memory for recent details and Long-Term Memory with redundancy-aware eviction, and further augment context-aware retrieval with Retrieval Alignment Contrastive Learning (RACL), achieving both high memory efficiency and strong temporal coherence.

    \item We demonstrate the effectiveness of WAT through extensive experiments, achieving state-of-the-art performance across multiple challenging online video understanding benchmarks and further constructing WAT-85K, a tailored training dataset for online VideoQA that explicitly addresses data scarcity in these tasks.
\end{itemize}

\section{Related Work}
\label{sec:relatedwork}

\subsection{MLLM for Video Understanding}
Multimodal Large Language Models (MLLMs)~\cite{ataallah2024minigpt4, bai2025qwen25vltechnicalreport, li2025videochat, li2024temporal, liu2024llavanext, ren2024timechat, zhang2025videollama, zhang2024internlm, zhang2024video, xu2024slowfast} have emerged as a dominant framework for video understanding, leveraging the reasoning capabilities of LLMs to process visual sequences. Early approaches followed a simple encode-and-reasoning pipeline: using a pretrained vision encoder to extract frame or clip features, which were then projected as visual tokens into the LLM for question answering or captioning. A primary challenge lies in long-video modeling, where MLLMs are often constrained by limited context windows and struggle with fine-grained spatiotemporal reasoning across dense frame sequences. To manage computational costs, a prevalent strategy is temporal sparsification and token compression. Methods like LLaVA-Video~\cite{zhang2024video} and SlowFast-LLaVA~\cite{xu2024slowfast} apply spatial-temporal pooling to reduce visual tokens. Video-LLaMA2~\cite{cheng2024videollama} further enhances this through hierarchical token aggregation. Beyond uniform compression, content-aware selection aims to identify and retain keyframes or shots. For instance, AKS~\cite{tang2025adaptive} uses CLIP~\cite{radford2021learning} similarity to select salient frames, while TSPO~\cite{tang2025tspo} leverages trainable parameters to select query-relevant events in videos. Chain-of-Shot~\cite{hu2025cos} employs a coarse-to-fine reasoning chain with off-the-shelf MLLMs to filter irrelevant shots. 

Existing Video-MLLMs are primarily designed for offline video understanding, where the full video or a pre-selected subset of frames is available before reasoning. Their core focus is temporal sparsification, or token compression, within a fixed context. In contrast, WAT explicitly targets online streaming scenarios and introduces a watching-before-thinking paradigm that decouples query-independent long-term visual perception from query-triggered reasoning.

\begin{figure*}[t]
\centering
\includegraphics[width=1.00\textwidth]{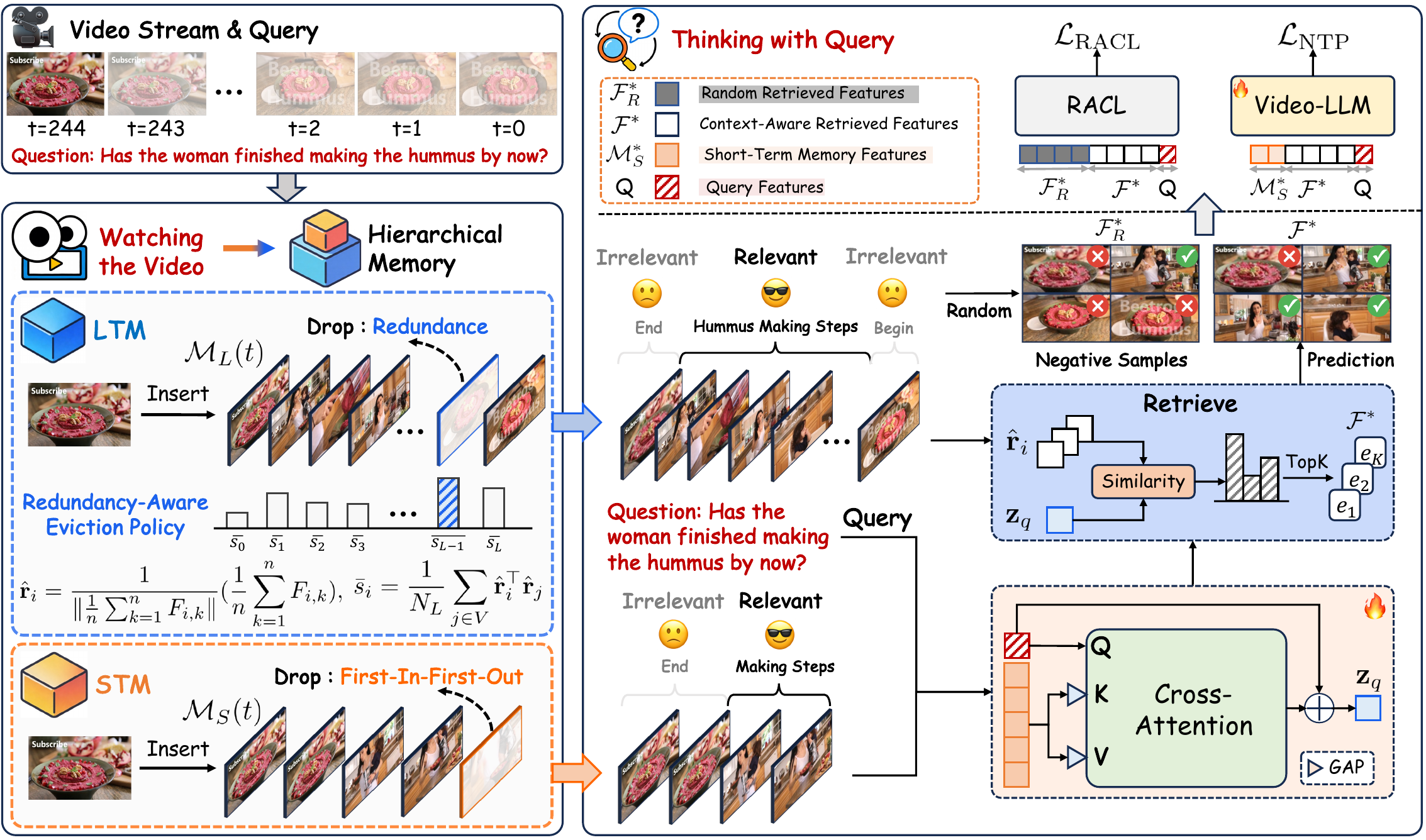}
  \caption{Overview of the proposed WAT framework, which follows a two-stage pipeline for online streaming video reasoning. Given a continuous video stream $V$ and a textual query $Q$, WAT decouples perception and reasoning into a Watching stage and a Thinking stage. In the Watching stage, WAT maintains a hierarchical memory system consisting of a Short-Term Memory (STM, $\mathcal{M}_S$) and a Long-Term Memory (LTM, $\mathcal{M}_L$). The STM serves as a high-fidelity sliding window, implemented as a First-In-First-Out (FIFO) queue to preserve recent frames and capture fine-grained temporal dynamics. In parallel, the LTM stores a compact yet semantically diverse set of historical features, managed by a redundancy-aware eviction policy that removes over-represented entries while protecting recent content. In the Thinking stage, triggered by a query $Q$, WAT performs context-aware retrieval by first fusing the query with the current STM content to form a conditioned query representation, which is then used to retrieve the top-$K$ most relevant historical features $\mathcal{F}^*$ from the LTM. Finally, the retrieved features $\mathcal{F}^*$, together with the short-term memory $\mathcal{M}_S$ and the query $Q$, are concatenated into a unified multimodal input sequence and fed into a Multimodal Large Language Model (MLLM) to perform coherent cross-temporal reasoning and generate the final response. The retrieval process is trained with an auxiliary Retrieval Alignment Contrastive Learning (RACL) objective. This contrastive loss aligns the relevant features $\mathcal{F}^*$ with the query $Q$ (positive) against random visual evidence $\mathcal{F}^*_R$ (negative).}
   \label{main}\
\end{figure*}

\subsection{MLLM for Online Video Understanding}
The task of online (streaming) video understanding, which requires processing continuously arriving video frames, was first formalized by VideoLLM-online~\cite{zhang2024internlm, chen2024videollm, di2025streaming, qian2024streaming, xu2025qwen2, zhang2024flash}. Subsequent research has evolved along two principal avenues to address its unique challenges. The first avenue focuses on efficient visual encoding and context management. A dominant strategy within this line employs dynamic memory banks to cache historical information, as seen in VideoStreaming~\cite{qian2024streaming}, Flash-VStream~\cite{zhang2024flash}, StreamChat~\cite{liu2024streamchat}, and VideoChat-online~\cite{huang2025online}. Alternative approaches optimize the underlying computational mechanism: ReKV~\cite{di2025streaming} and Inf-MLLM~\cite{ning2024inf} refine the management of the Key-Value (KV) cache for long video contexts, while VideoLLM-MoD~\cite{wu2024videollm} leverages Mixture-of-Depth (MoD) to dynamically reduce token count. A common characteristic of these methods is their reliance on linguistic guidance from user queries to identify and select semantically salient visual content. The second avenue explores the design of interaction paradigms. Works like MMDuet~\cite{wang2024videollm} refine the structural duet between video and text modalities, Dispider~\cite{qian2025dispider} introduces a novel disentangled framework based on a Perception-Decision-Reaction pipeline, and TimeChat-Online~\cite{yao2025timechat} relies on a Differential Token Dropping (DTD) design for efficiently encoding video streams.  

Prior online Video LLMs rely on query-driven or reactive memory updates, which may discard important visual information before a query is issued and thus hinder long-horizon reasoning. WAT instead adopts a proactive, query-agnostic watching stage with hierarchical memory to preserve semantically rich and diverse history, and only performs query-conditioned retrieval in a dedicated thinking stage. This separation enables more stable temporal coverage and efficient reasoning over long video streams.

\section{Method}
\label{sec:method}

\subsection{Overview}
A fundamental challenge in online video understanding is to achieve an optimal trade-off between \textit{memory efficiency} and \textit{contextual fidelity}—specifically, preserving fine-grained temporal details while maintaining long-term semantic coherence—under strict computational constraints. Existing approaches often struggle to resolve this tension. One common strategy employs aggressive temporal sampling or pruning to meet efficiency goals, but this inevitably severely compromises the model's capacity for spatiotemporal reasoning. Another line of work relies on small, real-time query-triggered memory updates; however, this purely reactive paradigm is prone to missing critical transient events.

To address this fundamental trade-off, we propose \textbf{WAT (Watching Before Thinking)}, a new two-stage online video understanding framework via hierarchical memory grounded in the principle that comprehensive visual perception must precede effective reasoning. This paradigm shift decouples immediate context preservation from long-term query-based semantic summarization, enabling robust and efficient online video understanding over extended streams.

Specifically, WAT processes a continuous video stream \(V = \{f_1, f_2, \dots, f_T\}\) and the accompanying textual queries \(Q\) through a structured pipeline:
\begin{itemize}
    \item \textbf{Watching Stage:} A hierarchical memory system maintains a dual-level representation of past frames. 
    \begin{itemize}
        \item \textit{Short-Term Memory (STM)} serves as a high-fidelity sliding-window buffer, preserving recent frames to capture fine-grained temporal dynamics and rich immediate context.
        \item \textit{Long-Term Memory (LTM)} serves as a compact yet diverse repository of historically salient frames, managed by a redundancy-aware eviction policy that ensures broad semantic coverage over long horizons.
    \end{itemize}
    \item \textbf{Thinking Stage:} Upon receiving a query \(Q\), a context-aware retrieval mechanism first fuses the query with STM content and then uses the enriched representation to retrieve the most relevant features from LTM. 
    \begin{itemize}
        \item \textit{Context-Aware Retrieval} implements Retrieval Alignment Contrastive Learning (RACL) to learn explicit context-level alignment constraints for context-aware retrieval effectively during training.
        \item \textit{Reasoning} integrates visual evidence, combining both recent and historical contexts, is finally fed into a Multimodal Large Language Model (MLLM) for coherent cross-temporal reasoning. 
    \end{itemize}  
\end{itemize}

This architecture not only addresses the twin challenges of memory efficiency and temporal coherence but also enables integration of real-time perception with long-horizon reasoning. In the following sections, we provide a detailed description of the design of both the watching and thinking components.

\subsection{Watching: Hierarchical Memory Architecture}

\subsubsection{Short-Term Memory}
The \textbf{Short-Term Memory (\(\mathcal{M}_S\))} component serves as a high-fidelity temporal buffer, implemented as a First-In-First-Out (FIFO) queue. This design ensures the model maintains immediate access to the most recent visual context while imposing minimal computational overhead. The STM's primary function is to preserve fine-grained temporal information essential for tasks requiring precise understanding of recent actions, object interactions, and dynamic scene evolution.



When the queue size exceeds its predetermined capacity, the oldest item is automatically evicted via dequeue operations. This continuous refresh mechanism ensures temporal continuity over short horizons while preventing memory overflow. 

\subsubsection{Long-Term Memory}

The \textbf{Long-Term Memory (\(\mathcal{M}_L\))} component addresses the critical challenge of retaining semantically diverse information across long video sequences while operating within strict memory constraints. With a fixed capacity \( N_L = 768 \), the LTM maintains a compact yet highly comprehensive representation of historically salient content, and effectively capturing the essential semantic evolution of the entire video sequence.

Each entry in the LTM comprises two representations to balance detailed reasoning with efficient memory management. The system preserves the original feature map \( R_i = F_t \) to support detailed spatiotemporal reasoning, while also maintaining a normalized descriptor \( \hat{\mathbf{r}}_i = \hat{\mathbf{R}}_i \) for efficient similarity computations and memory management. The descriptor is derived by applying global average pooling (GAP) across spatial dimensions, followed by L2 normalization:

\begin{equation}
   \mathbf{r}_i = \text{GAP}(F_i), \quad \hat{\mathbf{r}}_i = \frac{\mathbf{r}_i}{\|\mathbf{r}_i\|_2} 
\end{equation}

The cornerstone of LTM management is our innovative \textbf{redundancy-aware eviction policy}, which intelligently determines frame replacement when memory reaches capacity (\( |\mathcal{M}_L| = N_L \)). As shown in algorithm \ref{alg:ltm_update} This policy operates by computing pairwise cosine similarities between all stored descriptor vectors:
$S_{ij} = \hat{\mathbf{r}}_i^T \hat{\mathbf{r}}_j$.
The redundancy score for each frame \( i \) is then calculated as its average similarity to all other memory entries:

\begin{equation}
  \bar{s}_i = \frac{1}{N_L} \sum_{j=1}^{N_L} S_{ij}  
\end{equation}

To prevent premature deletion of novel or temporally critical content, the system implements a protection mechanism for the most recent \( \rho N_L \) frames (with protection ratio \( \rho = 0.1 \)). The eviction candidate is then selected as the frame with the highest redundancy score among the non-protected entries $i^* = \arg \max_{i \in V \setminus E} \bar{s}_i$ where \( E \) denotes the set of protected recent frames and \( V \) represents all current memory entries. This sophisticated eviction strategy ensures that the LTM maintains a rich and varied representation of the video's semantic evolution, systematically preventing over-representation of redundant visual content while promoting diversity across different temporal segments and scene types.

\begin{algorithm}[t]
\caption{Redundancy-Aware Eviction Policy}
\label{alg:ltm_update}
\begin{algorithmic}[1]
\Require new feature $F_t$, memory $(\mathcal{M}_L, \hat{R}, S, V)$, counters $(C, L)$, update freq $U$
\Ensure updated memory
\State $C \leftarrow C + 1$
\State $\hat{\mathbf{r}}_t \leftarrow \mathrm{normalize}(\mathrm{mean}(F_t))$
\If{$|\mathcal{M}_L| < N_L$}
    \State append $(F_t, \hat{\mathbf{r}}_t)$ to $\mathcal{M}_L$
\Else
    \If{$C - L \ge U$} 
        \State $S[V,V] \leftarrow \hat{R}[V]\hat{R}[V]^\top$; $L \leftarrow C$
    \EndIf
    \State $\bar{s}_i \leftarrow \mathrm{mean}(S[i,V])$ for $i \in V$
    \State $E \leftarrow$ most recent $\lceil 0.1|V| \rceil$ entries
    \State $i^* \leftarrow \arg\max_{i \in (V \setminus E)} \bar{s}_i$
    \State replace entry $i^*$ with $(F_t, \hat{\mathbf{r}}_t)$
\EndIf
\State update $S[i^*,:], S[:,i^*] \leftarrow \hat{\mathbf{r}}_t \hat{R}^\top$
\end{algorithmic}
\end{algorithm}

\subsubsection{Temporal Interaction of Memories}

WAT employs optimized update strategies for its Short-Term and Long-Term Memory components to address the challenge of maintaining temporal coherence in online video understanding. Their asynchronous operation enables efficient preservation of short-term temporal details and long-term semantic diversity under strict computational constraints.

The STM updates at a high frequency, continuously buffering incoming frames to capture fine-grained recent context. This design ensures responsiveness to dynamic scene changes and provides up-to-date visual information for accurate immediate reasoning.

In contrast, the LTM adopts a selective-replacement strategy governed by a redundancy-aware eviction policy. When reaching capacity, new frames are admitted while the most semantically redundant existing frame is evicted. Unlike periodic or event-driven updates in prior work, this strategy maintains a compact yet diverse long-term representation across the entire video, effectively reducing redundancy while preserving semantic diversity over extended temporal spans.

This decoupled architecture formally separates the high-frequency task of local frame buffering (handled by STM) from the low-frequency process of global semantic summarization (managed by LTM). The comprehensive system state at any time \( t \) is defined as the union of both memory components:

\begin{equation}
    \mathcal{M}(t) = \mathcal{M}_S(t) \cup \mathcal{M}_L(t)
\end{equation}

Each memory follows distinct update rules aligned with its temporal role. The asynchronous interaction between the STM and LTM enables WAT to preserve short-term temporal coherence and long-term semantic diversity under practical computational constraints. Moreover, the query-guided retrieval mechanism ensures that LTM features are both semantically relevant and temporally consistent with the current STM context, thereby supporting coherent cross-modal reasoning over long video sequences.

\begin{algorithm}[t]
\caption{Retrieval–Alignment Contrastive Learning}
\label{alg:racl_loss}
\begin{algorithmic}[1]
\Require question embeddings $\mathbf{Q}=\{\mathbf{q}_i\}_{i=1}^{B}$,
retrieved frame features $F^{*}=\{F_i^{*}\}_{i=1}^{B}$,
long-term memory $\mathcal{M}_L$,
temperature $\tau$
\Ensure contrastive loss $\mathcal{L}_{\text{RACL}}$

\State $\mathbf{r} \leftarrow \mathrm{normalize}\!\left(
        \frac{1}{B}\sum_{i=1}^{B} 
        \mathrm{mean}(F_i)
      \right)$

\State $\mathcal{N}_{\text{batch}} \leftarrow 
       \{\,\mathbf{r}^{(k)} \mid \mathbf{r}^{(k)} = \mathrm{permute}(\mathbf{r}, k),\;
       k=1,\dots,K\,\}$
\If{$|\mathcal{M}_L| > 0$}
    \State sample a subset $\{\tilde{F}_j\}_{j=1}^{M}$ from $\mathcal{M}_L$
    \State ${F_R^{*}} \leftarrow 
           \mathrm{normalize}\!\left(
           \frac{1}{M}\sum_{j=1}^{M} \mathrm{mean}(\tilde{F}_j)
           \right)$
    \State $\mathcal{N} \leftarrow \mathcal{N}_{\text{batch}} \cup \{\mathbf{r}_{\text{LTM}}\}$
\Else
    \State $\mathcal{N} \leftarrow \mathcal{N}_{\text{batch}}$
\EndIf

\For{$i=1$ to $B$}
    \State $s_i^{+} \leftarrow 
           \mathrm{sim}(\mathbf{q}_i, \mathbf{r}) / \tau$
    \State $s_{i,j}^{-} \leftarrow 
           \mathrm{sim}(\mathbf{q}_i, \mathbf{n}_j) / \tau,\;
           \forall \mathbf{n}_j \in \mathcal{N}$
    \State $\mathcal{L}_i \leftarrow 
           -\log \frac{\exp(s_i^{+})}
           {\exp(s_i^{+}) + \sum_j \exp(s_{i,j}^{-})}$
\EndFor

\State $\mathcal{L}_{\text{RACL}} \leftarrow \frac{1}{B}\sum_{i=1}^{B} \mathcal{L}_i$
\State \Return $\mathcal{L}_{\text{RACL}}$
\end{algorithmic}
\end{algorithm}

\subsection{Thinking: Context-Aware Retrieval and Reasoning}
\subsubsection{Context-Aware Retrieval}
To enable context-aware reasoning that seamlessly bridges immediate and historical visual content, WAT incorporates a sophisticated query-guided retrieval mechanism. This component dynamically selects the most relevant features from the LTM based on the synergistic integration of the current textual query and the visual context preserved within the STM.

Given an input query \( Q \) with its corresponding textual embedding \( \mathbf{Q} \in \mathbb{R}^d \), the system first performs multimodal fusion with the visual context from the STM to construct a context-aware query vector \( \mathbf{z_q} \). This fusion process is facilitated by an attention-based module \( \mathcal{A} \), which learns to optimally align textual query semantics with aggregated visual features from the STM: $\mathbf{z_q} = \mathbf{Q} + \mathcal{A}(\mathbf{Q}, \text{GAP}(\mathcal{M}_S))$

The resulting context-aware query vector \( \mathbf{z_q} \) subsequently serves as the basis for computing relevance scores across all entries within the LTM. This is achieved through cosine similarity measurements between the query vector and each normalized descriptor in long-term storage:

\begin{equation}
    s_i = \cos(\mathbf{z_q}, \hat{\mathbf{r}}_i) = \frac{\mathbf{z_q} \cdot \hat{\mathbf{r}}_i}{\|\mathbf{z_q}\| \|\hat{\mathbf{r}}_i\|}, \quad \forall i \in \{1, \ldots, |\mathcal{M}_L|\}
\end{equation}

The system then selects the top-\( K \) frames exhibiting the highest relevance scores:

\begin{equation}
    \mathcal{F}^* = \{e_i \mid i \in \arg \text{TopK}(\{s_i\}, K), e_i \in \mathcal{M}_L\}
\end{equation}

The retrieved features \( \mathcal{F}^* \) are concatenated with the complete content of the STM to form the comprehensive visual evidence \( \mathcal{F} \) for subsequent MLLM processing
$\mathcal{F} = \mathcal{M}_S \parallel \mathcal{F}^*$.
This retrieval mechanism ensures that the final reasoning process is informed by both the fine-grained recent context from the STM and the semantically relevant, long-term context selectively retrieved from the LTM, enabling robust cross-temporal reasoning across diverse video understanding tasks.

\input{table/streamingbench}

\subsubsection{Retrieval Alignment Contrastive Learning}
Beyond the vanilla MLLM training pipeline, we employ Retrieval Alignment Contrastive Learning (RACL) as an auxiliary, self-supervised objective. RACL enhances the model's ability to retrieve query-relevant semantics from the long-term memory without requiring additional annotations.

To supervise alignment between query representations and WAT's memory-aware retrieval,we formulate RACL as a contrastive learning objective. This objective encourages retrieved visual evidence to be semantically consistent with the current question while remaining discriminative against redundant or temporally irrelevant memory content.

As shown in algorithm \ref{alg:racl_loss}, given a batch of question embeddings $\mathbf{Q}$, WAT retrieves frame-level features $\mathcal{M}_S$ and $\mathcal{M}_L$ from its Working Memory (STM/LTM) for each query.
We aggregate the retrieved features via temporal averaging followed by batch-wise pooling to calculate a global retrieval representation $\mathbf{r}$ by $\frac{1}{B} \sum_{i=1}^{B} \left( \frac{1}{T} \sum_{t=1}^{T} f_{i,t} \right)$ which serves as the positive retrieval anchor.

To construct informative negatives, RACL combines in-batch shifted retrieval representations and hard negatives sampled $\mathbf{n}_j$ from random Long-Term Memory samples $F_R^*$. The latter introduces temporally misaligned distractors, reflecting the persistence characteristics of WAT's memory. For each question $\mathbf{q}_i$, we compute temperature-scaled cosine similarities:
\begin{equation}
s_i^{+} = \frac{\cos(\mathbf{q}_i, \mathbf{r})}{\tau}, \quad
s_{i,j}^{-} = \frac{\cos(\mathbf{q}_i, \mathbf{n}_j)}{\tau},
\label{eq:racl_sim}
\end{equation}
where $\tau$ is temperature, then optimize an InfoNCE objective:
\begin{equation}
\mathcal{L}_{\text{RACL}}
= -\frac{1}{B} \sum_{i=1}^{B}
\log
\frac{\exp(s_i^{+})}
{\exp(s_i^{+}) + \sum_j \exp(s_{i,j}^{-})}.
\label{eq:racl_loss}
\end{equation}

By grounding contrastive supervision on WAT's retrieval outputs and memory structure, RACL improves query-evidence alignment.

\input{table/ovobench}

\subsubsection{MLLM Reasoning and Inference}
Our training objective combines the aforementioned RACL loss with the standard MLLM next-token prediction loss, formally expressed as $\mathcal{L} = \mathcal{L}{\text{RACL}} + \mathcal{L}{\text{NTP}}$, thereby jointly optimizing both the retrieval and reasoning capabilities of the model. This joint optimization allows the system to learn semantically aligned representations while preserving temporal coherence across long video sequences. During inference, only the MLLM reasoning path is retained, ensuring efficient online operation. The trained retriever and LLM operate in sequence, with the retriever first identifying the most relevant visual context based on the query, and the LLM subsequently generating the answer by integrating the retrieved context with the query information, producing accurate and temporally coherent responses.

\subsection{WAT-85K Collection}
To support effective training of online Video LLMs, we introduce WAT-85K, a synthetic dataset tailored for online VideoQA. The dataset is constructed from two primary sources, with rigorous filtering and augmentation to ensure quality and relevance for online reasoning.
\textbf{Offline Data.} We first curated a subset from LLaVA-Video-178K~\cite{zhang2024video}. Since the original collection inherently lacks genuine long-form videos, our selection strategy prioritized a balanced mixture of short- and medium-length videos to foster generalizable temporal understanding.
\textbf{Online Data.} The primary source for this segment is TimeChat-Online-178K~\cite{yao2025timechat}. We identified that this dataset contained a substantial number of short videos and clips extracted from longer sequences. To better align with the continuous nature of streaming video, we filtered out most short videos and, for videos available in clips, retained the complete long-form source videos instead of the fragmented segments. Furthermore, we observed significant scarcity and quality issues in the existing data for the \textit{real-time perception} and \textit{proactive forecasting} tasks. To address this gap, we manually constructed a substantial number of high-quality samples for these critical online reasoning scenarios. 

\section{Experiment}
\label{sec:experiment}

\subsection{Implementation Details}
We construct our model based on the Qwen2.5VL-7B~\cite{bai2025qwen25vltechnicalreport} architecture. \textbf{For training}, we sample frames at 1 FPS with a maximum video length of 4000 seconds for online VideoQA samples. We configure the model with a input resolution of $448\times448$ pixels, short-term memory capacity $N_{L}$=16, long-term memory capacity $N_{L}$=768, the number of retrieved features $N_{\mathcal{F}^*}$=32, update frequency $U$=64, batch size 16, and learning rate 1e-5. Our training dataset WAT-85K combines both offline and online data. We keep the vision encoder frozen while fine-tuning the WAT module, full-parameter projector, and language model for 1 epoch. All experiments are performed on 8$\times$A800 80G GPUs. \textbf{During inference}, we set 1 FPS for all benchmarks' frame processing rate.

\subsection{Results on Online Video Benchmarks}
We evaluate WAT on two established online VideoQA benchmarks:StreamingBench~\cite{lin2024streamingbench} and OVO-Bench~\cite{niu2025ovo}. For these evaluations, VideoLLMs processes historical video content up to the current timestamp, which represents the moment when a user question is posed.

\subsubsection{StreamingBench}
We evaluate WAT on StreamingBench, which focuses on Real-Time Visual Understanding across ten diverse tasks, including Object Perception (OP), Causal Reasoning (CR), Clips Summarization (CS), Attribute Perception (ATP), Event Understanding (EU), Text-Rich Understanding (TR), Prospective Reasoning (PR), Spatial Understanding (SU), Action Perception (ACP), and Counting (CT). As reported in Table~\ref{tab:streamingbench}, WAT achieves the best overall performance (77.70) among all open-source online Video LLMs, establishing a new state-of-the-art under online conditions.
Compared to recent SOTA baselines such as TimeChat-Online-7B~\cite{yao2025timechat} (75.36) and Dispider-7B~\cite{qian2025dispider} (67.63), WAT consistently improves across nearly all categories, particularly in OP (82.93), CS (85.49), ATP (84.24), EU (76.58), TR (83.80), PR(83.33), SU (73.17) and ACP (72.44). These tasks demand both fine-grained temporal reasoning and strong contextual awareness, thereby demonstrating that WAT’s hierarchical temporal memory can maintain coherent representations over extended video streams and accurately retrieve relevant visual information from the hierarchical memory.
Notably, despite operating under the same real-time constraint (1fps), WAT also surpasses the proprietary large-scale systems such as Gemini 1.5 Pro~\cite{team2024gemini} (75.69) and GPT-4o~\cite{hurst2024gpt} (73.28), while using only open-source backbones and online inference. This result confirms that the proposed Watching-Before-Thinking paradigm effectively balances efficiency and long-horizon reasoning, enabling robust real-time perception without the need for offline full-context encoding.
Overall, these findings highlight WAT’s superior adaptability to online scenarios, outperforming all existing open-source online models and approaching the performance of proprietary multimodal LLMs.

\subsubsection{OVO-Bench}
We further evaluate WAT on the OVO-Bench, which assesses comprehensive video understanding capabilities across three aspects: Real-Time Visual Perception, Backward Tracing, and Forward Active Responding. As presented in Table~\ref{tab:ovo-bench}, our method achieves an overall average score of 55.2, establishing a new state-of-the-art among open-source online Video LLMs.
Compared with existing online baselines such as TimeChat-Online-7B (46.7) and Flash-VStream-7B~\cite{zhang2024flash} (33.2), WAT shows consistent improvements across all categories. In the Real-Time Visual Perception group, WAT achieves an average of 64.7, outperforming prior online systems by a large margin while surpassing the accuracy of offline counterparts. This demonstrates that the proposed Watching-Before-Thinking mechanism effectively captures temporal dependencies even under sparse frame sampling (1fps).
For Backward Tracing, WAT obtains 45.2 surpassing TimeChat-Online-7B (41.7) and significantly outperforming other online models that lack long-term temporal modeling. This improvement arises from our hierarchical memory architecture, which enables efficient recall of past observations without full re-encoding. In the Forward Active Responding tasks, WAT reaches 55.8, showing clear advantages in multi-step reasoning and sequential understanding. Notably, WAT maintains both temporal awareness and causal consistency across long-horizon sequences, which are typically challenging for existing online models. While proprietary multimodal systems such as Gemini 1.5 Pro and GPT-4o still achieve higher overall scores due to their large-scale proprietary training, WAT significantly narrows the performance gap in a fully open-source, real-time setting. These results clearly validate that our design effectively bridges the performance-efficiency trade-off for online video understanding.

\input{table/offline}
\input{table/ab_memory}
\input{table/ab_ltm}
\input{table/ab_racl}

\subsection{Results on Offline Video Benchmarks}
Although WAT is primarily designed for online video understanding, it also demonstrates strong generalization ability to offline long video benchmarks. As shown in Table~\ref{tab:long_video_benchmarks}, WAT achieves 63.9 accuracy on MLVU~\cite{zhou2024mlvu}, as well as 62.4 and 50.8 on VideoMME~\cite{fu2025video} in the overall and long-video subsets, respectively, using only 1 fps frame sampling.
Compared with existing open-source offline VideoLLMs, many of which rely on substantially denser frame sampling (e.g., 64–2048 frames) or larger context windows to capture long-range dependencies, WAT achieves competitive or superior performance under a significantly more constrained input budget. Notably, WAT also performs on par with, or even surpasses, several recent online VideoLLMs specifically designed for streaming scenarios, while being evaluated here in a fully online setting.
These results suggest that the proposed context-aware retrieval and hierarchical memory design enable WAT to effectively summarize long-term video dynamics without densely observing every frame. Beyond improving efficiency in streaming scenarios, these design choices also preserve robust long-range reasoning capability when applied to the offline setting, highlighting the generality and scalability of the proposed framework.

\begin{figure*}[t]
\centering
\includegraphics[width=0.99\textwidth]{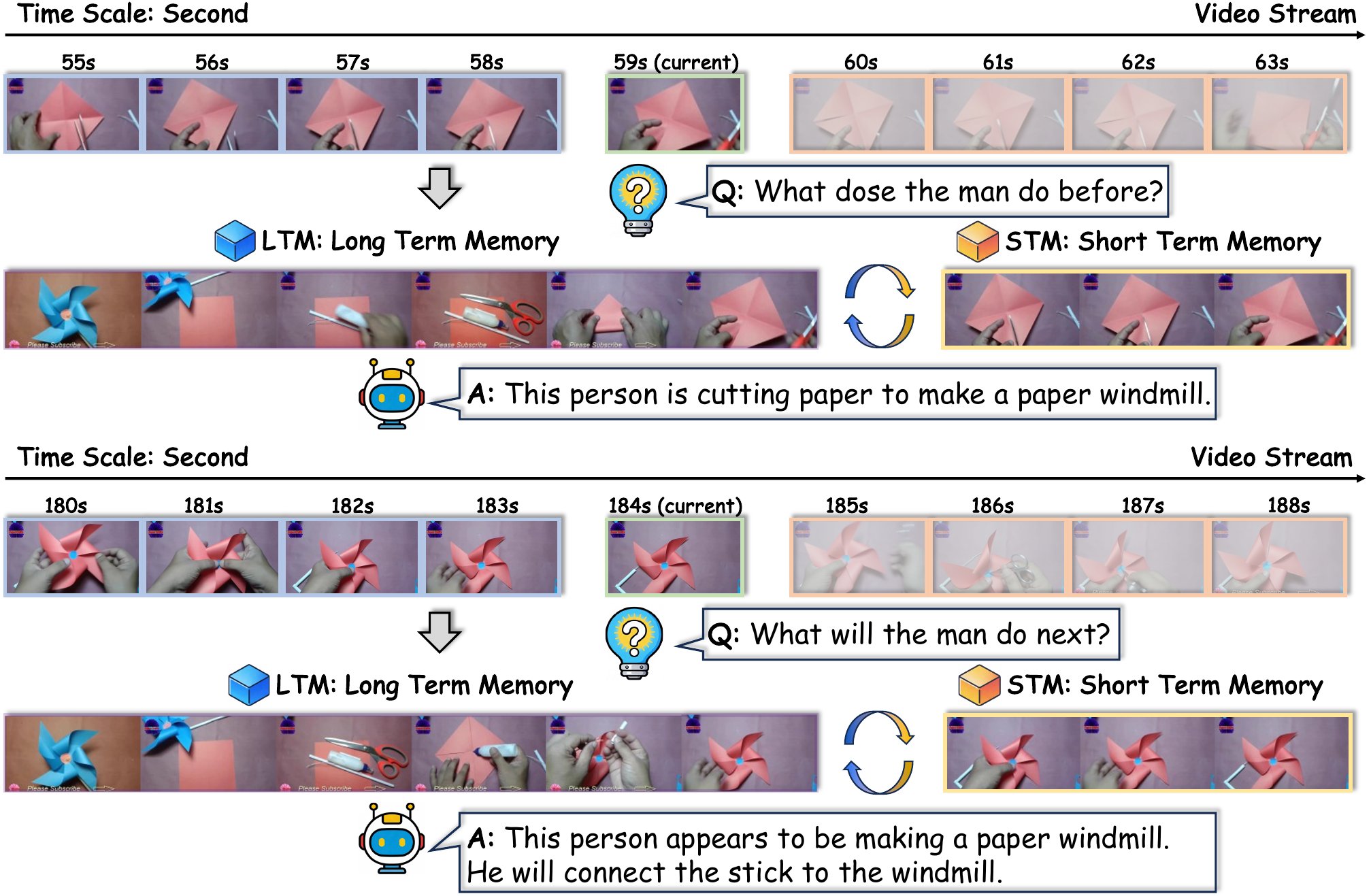}
  \caption{Case study of WAT on a sample video from OVO-Bench demonstrates its online reasoning capabilities. At 59s, when a user poses the query “What does the man do before?”, WAT proactively infers the preceding actions and correctly outputs the past activity, showing its ability to leverage hierarchical memory for temporal context. Furthermore, at 184s, when presented with a follow-up question about the future, “What will the man do next?”, the system similarly produces an accurate prediction, highlighting WAT’s capacity for both backward tracing and forward forecasting in continuous video streams.}
   \label{exp}\
\end{figure*}

\subsection{Ablation Study}
\subsubsection{Hierarchical Memory Design}
Table \ref{tab:memory}\ reports the ablation results of the proposed hierarchical memory system. Compared with the base model Qwen2.5-VL, introducing only long-term memory (LTM) improves performance on StreamingBench from 73.7 to 75.7, indicating the benefit of retaining global contextual information. Further incorporating short-term memory (STM) yields consistent gains on both StreamingBench (+2.0) and OVO-Bench (+4.9), highlighting the importance of modeling recent temporal context in streaming scenarios. These results demonstrate that STM and LTM are complementary, and their combination yields the best overall performance.

\subsubsection{LTM Length}
As shown in Table \ref{tab:ltm}, the impact of different long-term memory sizes. Increasing LTM length consistently improves performance on both benchmarks, with more pronounced gains on OVO-Bench. In particular, expanding $N_{\mathrm{LTM}}$ from 32 to 768 yields a +3.2 improvement on OVO-Bench, indicating that larger memory capacity better preserves long-range visual information. In contrast, StreamingBench shows more moderate gains, suggesting diminishing returns when recent context dominates the task.

\begin{figure*}[t]
\centering
\includegraphics[width=0.99\textwidth]{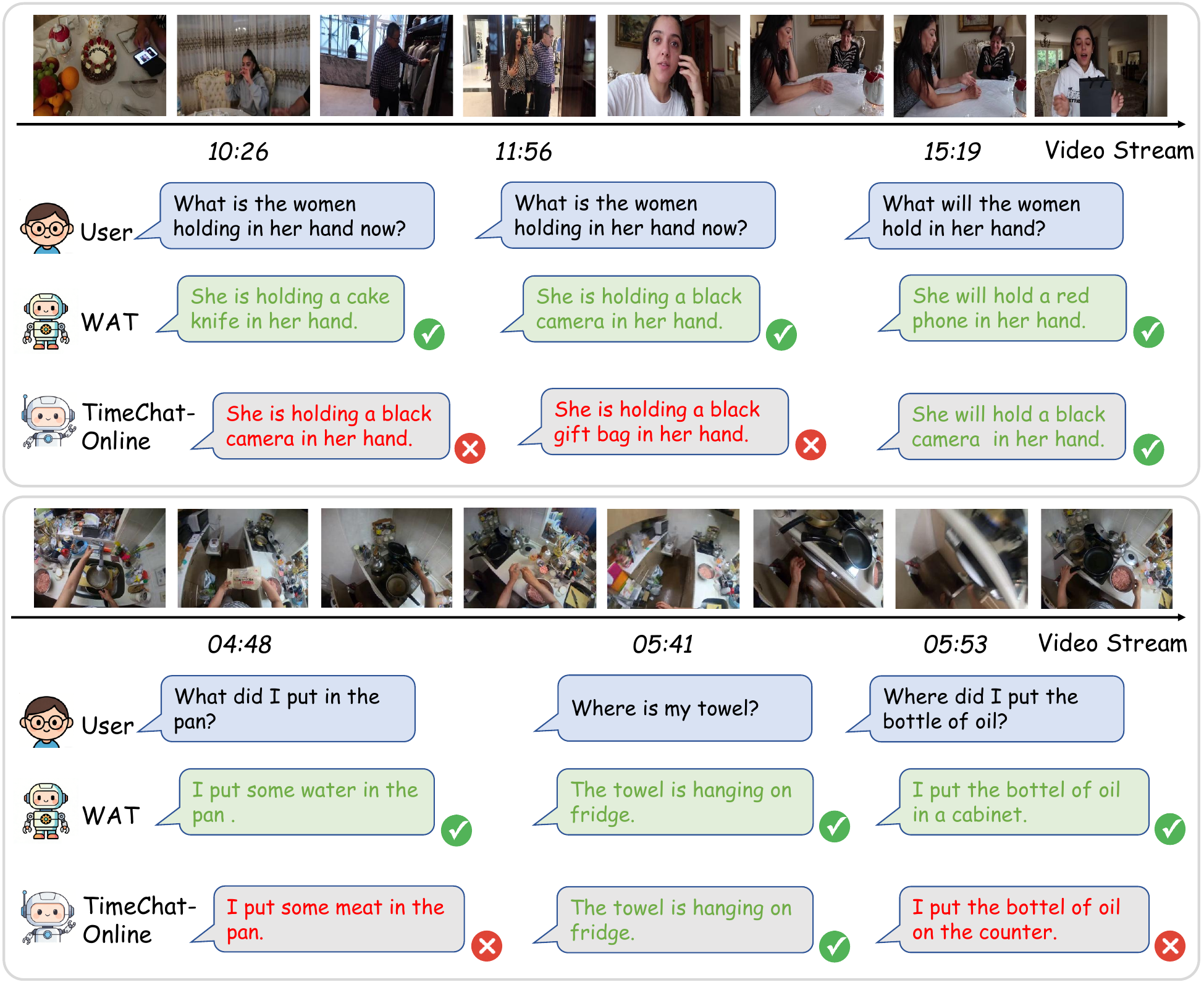}
  \caption{Visualization comparisons of retrieved visual content and corresponding responses between our method and other state-of-the-art (SOTA) approaches. In the top example, WAT exhibits strong capabilities in long-term temporal reasoning, effectively capturing dependencies across distant frames for online video understanding. This highlights WAT’s advantage in modeling extended temporal contexts, which is crucial for accurate understanding of complex, continuous video streams. The bottom example further demonstrates WAT's robust performance on conventional (short-form) VideoQA tasks, showing precise alignment between relevant visual segments and generated responses.}
   \label{exp2}\
\end{figure*}

\subsubsection{Loss Construction}
Table \ref{tab:racl} presents an ablation study on the loss construction. Using only the next-token prediction loss $\mathcal{L}_{\text{NTP}}$ provides a strong baseline, achieving 50.1 on OVO-Bench and 76.4 on StreamingBench. Introducing the proposed retrieval–alignment contrastive loss $\mathcal{L}_{\text{RACL}}$ leads to significant improvements across both OVO-Bench(from 50.1 to 54.6) and StreamingBench(from 76.4 to 77.3), demonstrating the effectiveness of explicitly aligning question representations with retrieved visual evidence.
We further examine the effect of the temperature parameter $\tau$ in $\mathcal{L}_{\text{RACL}}$. Increasing $\tau$ from 0.01 to 0.07 yields additional gains, improving OVO-Bench performance from 54.6 to 55.2 and StreamingBench from 77.3 to 77.7. This suggests that a moderately relaxed contrastive distribution facilitates more stable optimization and better feature alignment. Overall, these results indicate that $\mathcal{L}_{\text{RACL}}$ complements $\mathcal{L}_{\text{NTP}}$, and their combination results in more discriminative representations and improved performance in both offline and streaming evaluation settings.

\subsubsection{Qualitative Results}
Figure \ref{exp} provides a comprehensive visualization of the WAT framework's two-stage pipeline (Watching and Thinking) alongside a representative example. The workflow begins as WAT watches the uninterrupted video stream: each frame is encoded, and its features are dynamically enqueued into the hierarchical memory system—where the Short-Term Memory (STM) maintains recent context and the Long-Term Memory (LTM) selectively consolidates diverse historical semantics through its redundancy-aware eviction policy. Upon receiving a query, the framework transitions to the thinking stage: it actively retrieves the most relevant frames from the LTM by fusing the query with the current STM context, and finally synthesizes this multimodal evidence to generate the accurate, coherent answer.

As evidenced in Figure~\ref{exp2}, compared to the SOTA online Video LLMs, WAT achieves superior performance on both temporal reasoning and conventional VideoQA tasks. The Upper example further reveal its stronger capability in event and temporal logic understanding. Notably, WAT exhibits more pronounced advantages in scenarios requiring robust long-range temporal dependency modeling, while also delivering steady consistent gains on standard VideoQA benchmarks. The improvements across these two categories indicate that WAT effectively enhances fine-grained temporal understanding without compromising general video comprehension capability. These results substantiate the efficacy of the proposed watching-before-thinking framework, demonstrating its robustness in online video understanding scenarios.

\section{Conclusion}
This paper presents WAT, a new two-stage framework for online video understanding. WAT addresses the fundamental tension between memory efficiency and long-horizon temporal reasoning in streaming scenarios by explicitly decoupling query-independent visual perception from query-triggered reasoning. Specifically, we introduce a two-stage pipeline that integrates a high-fidelity Short-Term Memory with a redundancy-aware Long-Term Memory, together with a context-aware retrieval mechanism optimized by Retrieval Alignment Contrastive Learning, enabling temporally coherent and query-relevant reasoning over extended video streams. To facilitate effective training in genuine online settings, we further construct WAT-85K, a large-scale dataset that emphasizes real-time perception, backward tracing, and proactive forecasting, covering a diverse range of challenging streaming video scenarios. Extensive experiments on multiple online and offline benchmarks demonstrate that WAT consistently outperforms existing online Video LLMs and remains competitive with strong offline models even under strict streaming constraints. These results validate not only the effectiveness and efficiency of the proposed framework, but also its robustness and practicality for real-world long video understanding applications.

\vfill

\end{document}

%% file: table/streamingbench.tex
\begin{table*}[t]
    \small
    \centering
    \caption{Performance comparison on StreamingBench focusing on \textit{Real-Time Visual Understanding} tasks. Real-Time Visual Understanding encompasses Object Perception (OP), Causal Reasoning (CR), Clips Summarization (CS), Attribute Perception (ATP), Event Understanding (EU), Text-Rich Understanding (TR), Prospective Reasoning (PR), Spatial Understanding (SU), Action Perception (ACP), and Counting (CT). The \textbf{bold} values indicate the best performance, and \uline{underlined} values indicate the second best.}
    \label{tab:streamingbench}
    \small
\adjustbox{width=\linewidth}
  {
  \setlength{\tabcolsep}{7pt}
    \begin{tabular}{l@{\hspace{3pt}}c@{\hspace{3pt}} | c c c c c c c c c c | c}
    \toprule
    \multirow{1}{*}{\textbf{Model}} & \multirow{1}{*}{\textbf{\# Frames}} & \textbf{OP} & \textbf{CR} & \textbf{CS} & \textbf{ATP} & \textbf{EU} & \textbf{TR} & \textbf{PR} & \textbf{SU} & \textbf{ACP} & \textbf{CT} & \textbf{All} \\
    \midrule
    Human & - 
        & 89.47 & 92.00 & 93.60 & 91.47 & 95.65 & 92.52 & 88.00 & 88.75 & 89.74 & 91.30 & 91.46 \\
    \midrule
    \multicolumn{13}{c}{\textbf{Proprietary MLLMs}} \\
    \midrule
    Gemini 1.5 pro~\cite{team2024gemini} & 1 fps 
        & 79.02 & 80.47 & 83.54 & 79.67 & 80.00 & 84.74 & 77.78 & 64.23 & 71.95 & 48.70 & 75.69 \\
    GPT-4o~\cite{hurst2024gpt} & 64 
        & 77.11 & 80.47 & 83.91 & 76.47 & 70.19 & 83.80 & 66.67 & 62.19 & 69.12 & 49.22 & 73.28 \\
    Claude 3.5 Sonnet~\cite{anthropic2024claude35sonnet} & 20 
        & 73.33 & 80.47 & 84.09 & 82.02 & 75.39 & 79.53 & 61.11 & 61.79 & 69.32 & 43.09 & 72.44 \\
    \midrule
    \multicolumn{13}{c}{\textbf{Open-source Offline VideoLLMs}} \\
    \midrule
    \textcolor{grey}{Video-LLaMA2-7B~\cite{cheng2024videollama}} & \textcolor{grey}{32} 
        & \textcolor{grey}{55.86} & \textcolor{grey}{55.47} & \textcolor{grey}{57.41} & \textcolor{grey}{58.17} & \textcolor{grey}{52.80} & \textcolor{grey}{43.61} & \textcolor{grey}{39.81} & \textcolor{grey}{42.68} & \textcolor{grey}{45.61} & \textcolor{grey}{35.23} & \textcolor{grey}{49.52} \\
    \textcolor{grey}{VILA-1.5-8B~\cite{lin2023vila}} & \textcolor{grey}{14}
        & \textcolor{grey}{53.68} & \textcolor{grey}{49.22} & \textcolor{grey}{70.98} & \textcolor{grey}{56.86} & \textcolor{grey}{53.42} & \textcolor{grey}{53.89} & \textcolor{grey}{54.63} & \textcolor{grey}{48.78} & \textcolor{grey}{50.14} & \textcolor{grey}{17.62} & \textcolor{grey}{52.32} \\
    \textcolor{grey}{Video-CCAM-14B~\cite{fei2024video}} & \textcolor{grey}{96} 
        & \textcolor{grey}{56.40} & \textcolor{grey}{57.81} & \textcolor{grey}{65.30} & \textcolor{grey}{62.75} & \textcolor{grey}{64.60} & \textcolor{grey}{51.40} & \textcolor{grey}{42.59} & \textcolor{grey}{47.97} & \textcolor{grey}{49.58} & \textcolor{grey}{31.61} & \textcolor{grey}{53.96} \\
    \textcolor{grey}{LongVA-7B~\cite{zhang2024longva}} & \textcolor{grey}{128}
        & \textcolor{grey}{70.03} & \textcolor{grey}{63.28} & \textcolor{grey}{61.20} & \textcolor{grey}{70.92} & \textcolor{grey}{62.73} & \textcolor{grey}{59.50} & \textcolor{grey}{61.11} & \textcolor{grey}{53.66} & \textcolor{grey}{54.67} & \textcolor{grey}{34.72} & \textcolor{grey}{59.96} \\
    \textcolor{grey}{InternVL-V2-8B~\cite{chen2024internvl}} & \textcolor{grey}{16}
        & \textcolor{grey}{68.12} & \textcolor{grey}{60.94} & \textcolor{grey}{69.40} & \textcolor{grey}{77.12} & \textcolor{grey}{67.70} & \textcolor{grey}{62.93} & \textcolor{grey}{59.26} & \textcolor{grey}{53.25} & \textcolor{grey}{54.96} & \textcolor{grey}{56.48} & \textcolor{grey}{63.72} \\
    \textcolor{grey}{Kangaroo-7B~\cite{kangaroogroup}} & \textcolor{grey}{64}
        & \textcolor{grey}{71.12} & \textcolor{grey}{84.38} & \textcolor{grey}{70.66} & \textcolor{grey}{73.20} & \textcolor{grey}{67.08} & \textcolor{grey}{61.68} & \textcolor{grey}{56.48} & \textcolor{grey}{55.69} & \textcolor{grey}{62.04} & \textcolor{grey}{38.86} & \textcolor{grey}{64.60} \\
    \textcolor{grey}{LLaVA-NeXT-Video-32B~\cite{zhang2024video}} & \textcolor{grey}{64}
        & \textcolor{grey}{78.20} & \textcolor{grey}{70.31} & \textcolor{grey}{73.82} & \textcolor{grey}{76.80} & \textcolor{grey}{63.35} & \textcolor{grey}{69.78} & \textcolor{grey}{57.41} & \textcolor{grey}{56.10} & \textcolor{grey}{64.31} & \textcolor{grey}{38.86} & \textcolor{grey}{66.96} \\
    \textcolor{grey}{MiniCPM-V-2.6-8B~\cite{yao2024minicpm}} & \textcolor{grey}{32} 
        & \textcolor{grey}{71.93} & \textcolor{grey}{71.09} & \textcolor{grey}{77.92} & \textcolor{grey}{75.82} & \textcolor{grey}{64.60} & \textcolor{grey}{65.73} & \textcolor{grey}{70.37} & \textcolor{grey}{56.10} & \textcolor{grey}{62.32} & \textcolor{grey}{53.37} & \textcolor{grey}{67.44} \\
    \textcolor{grey}{LLaVA-OneVision-7B~\cite{li2024llava}} & \textcolor{grey}{32} 
        & \textcolor{grey}{80.38} & \textcolor{grey}{74.22} & \textcolor{grey}{76.03} & \textcolor{grey}{80.72} & \textcolor{grey}{72.67} & \textcolor{grey}{71.65} & \textcolor{grey}{67.59} & \textcolor{grey}{65.45} & \textcolor{grey}{65.72} & \textcolor{grey}{45.08} & \textcolor{grey}{71.12} \\
    \textcolor{grey}{Qwen2.5-VL-7B~\cite{xu2025qwen2}} & \textcolor{grey}{1 fps}
        & \textcolor{grey}{78.32} & \textcolor{grey}{80.47} & \textcolor{grey}{78.86} & \textcolor{grey}{80.45} & \textcolor{grey}{76.73} & \textcolor{grey}{78.50} & \textcolor{grey}{79.63} & \textcolor{grey}{63.41} & \textcolor{grey}{66.19} & \textcolor{grey}{53.19} & \textcolor{grey}{73.68} \\
    \addlinespace[1pt]
    \midrule
    \multicolumn{13}{c}{\textbf{Open-source  Online VideoLLMs}} \\
    \midrule
    Flash-VStream-7B~\cite{zhang2024flash}~\textsubscript{\textcolor{grey}{[ArXiv 2024]}} & 1 fps
        & 25.89 & 43.57 & 24.91 & 23.87 & 27.33 & 13.08 & 18.52 & 25.20 & 23.87 & \uline{48.70} & 23.23 \\
    VideoLLM-online-8B~\cite{chen2024videollm}~\textsubscript{\textcolor{grey}{[CVPR 2024]}} & 2 fps
        & 39.07 & 40.06 & 34.49 & 31.05 & 45.96 & 32.40 & 31.48 & 34.16 & 42.49 & 27.89 & 35.99 \\
    Dispider-7B~\cite{qian2025dispider}~\textsubscript{\textcolor{grey}{[CVPR 2025]}} & 1 fps
        & 74.92 & 75.53 & 74.10 & 73.08 & 74.44 & 59.92 & 76.14 & 62.91 & 62.16 & 45.80 & 67.63 \\
    TimeChat-Online-7B~\cite{yao2025timechat}~\textsubscript{\textcolor{grey}{[MM 2025]}} &1 fps
        & \uline{80.22} & \textbf{82.03} & \uline{79.50} & \uline{83.33} & \uline{76.10} & \uline{78.50} & \uline{78.70} & \uline{64.63} & \uline{69.60} & \textbf{57.98} & \uline{75.36} \\
    \midrule
    \rowcolor{blue!8}
    \textbf{WAT} & 1 fps
        & \textbf{82.93} & \uline{78.91} & \textbf{85.49} & \textbf{84.24} & \textbf{76.58} & \textbf{83.80} & \textbf{83.33} & \textbf{73.17} & \textbf{72.44} & 45.74 & \textbf{77.70} \\
    \bottomrule
    \end{tabular}
}
\end{table*}

%% file: table/ovobench.tex
\begin{table*}[t]
    \centering
    \caption{Evaluation results on OVO-Bench comprising three categories: i) \textit{Real-Time Visual Perception} (OCR: Optical Character Recognition, ACR: Action Recognition, ATR: Attribute Recognition, STU: Spatial Understanding, FPD: Future Prediction, OJR: Object Recognition), ii) \textit{Backward Tracing} (EPM: Episodic Memory, ASI: Action Sequence Identification, HLD: Hallucination Detection), and iii) \textit{Forward Active Responding} (REC: Repetition Event Count, SSR: Sequential Steps Recognition, CRR: Clues Reveal Responding). The \textbf{bold} values indicate the best performance, and \uline{underlined} values indicate the second best.}
    \label{tab:ovo-bench}

\adjustbox{width=\linewidth}
  {
  \setlength{\tabcolsep}{2pt}
    \begin{tabular}{l@{}c@{\hspace{3pt}}|ccccccc|cccc|>{\centering\arraybackslash}p{0.75cm}>{\centering\arraybackslash}p{0.75cm}>{\centering\arraybackslash}p{0.75cm}>{\centering\arraybackslash}p{0.75cm}|c}
    \toprule
    \multirow{2}{*}{\textbf{Model}} & \multirow{2}{*}{\textbf{\#Frames}} & 
    \multicolumn{7}{c|}{\textbf{Real-Time Visual Perception}} & 
    \multicolumn{4}{c|}{\textbf{Backward Tracing}} & 
    \multicolumn{4}{c|}{\textbf{Forward Active Responding}} & \multirow{2}{*}{\textbf{Overall}} \\
    \addlinespace[2pt]
    \cmidrule[0.5pt](lr){3-9} \cmidrule[0.5pt](lr){10-13} \cmidrule[0.5pt](lr){14-17}
    \addlinespace[2pt]
    &  & OCR & ACR & ATR & STU & FPD & OJR & Avg. & 
    EPM & ASI & HLD & Avg. & 
    REC & SSR & CRR & Avg. & Avg. \\
    \midrule
    Human & - & 94.0 & 92.6 & 94.8 & 92.7 & 91.1 & 94.0 & 93.2 & 92.6 & 93.0 & 91.4 & 92.3 & 95.5 & 89.7 & 93.6 & 92.9 & 92.8 \\
    \midrule
    \multicolumn{18}{c}{\textbf{Proprietary Multimodal Models}} \\
    \midrule
    Gemini 1.5 Pro~\cite{team2024gemini} & 1fps & 87.3 & 67.0 & 80.2 & 54.5 & 68.3 & 67.4 & 70.8 & 68.6 & 75.7 & 52.7 & 62.3 & 35.5 & 74.2 & 61.7 & 57.2 & 65.3 \\
    GPT-4o~\cite{hurst2024gpt} & 64 & 69.1 & 65.1 & 65.5 & 50.0 & 68.3 & 63.7 & 63.6 & 49.8 & 71.0 & 55.4 & 58.7 & 27.6 & 73.2 & 59.4 & 53.4 & 58.6 \\
    \midrule
    \multicolumn{18}{c}{\textbf{Open-source Offline VideoLLMs}} \\
    \midrule
    \textcolor{grey}{LLaVA-NeXT-Video-7B~\cite{zhang2024video}} & \textcolor{grey}{64} & \textcolor{grey}{69.8} & \textcolor{grey}{59.6} & \textcolor{grey}{66.4} & \textcolor{grey}{50.6} & \textcolor{grey}{72.3} & \textcolor{grey}{61.4} & \textcolor{grey}{63.3} & \textcolor{grey}{51.2} & \textcolor{grey}{64.2} & \textcolor{grey}{9.7} & \textcolor{grey}{41.7} & \textcolor{grey}{34.1} & \textcolor{grey}{67.6} & \textcolor{grey}{60.8} & \textcolor{grey}{54.2} & \textcolor{grey}{53.1} \\
    \textcolor{grey}{LLaVA-OneVision-7B~\cite{li2024llava}} & \textcolor{grey}{64} & \textcolor{grey}{67.1} & \textcolor{grey}{58.7} & \textcolor{grey}{69.8} & \textcolor{grey}{49.4} & \textcolor{grey}{71.3} & \textcolor{grey}{60.3} & \textcolor{grey}{62.8} & \textcolor{grey}{52.5} & \textcolor{grey}{58.8} & \textcolor{grey}{23.7} & \textcolor{grey}{45.0} & \textcolor{grey}{24.8} & \textcolor{grey}{66.9} & \textcolor{grey}{60.8} & \textcolor{grey}{50.9} & \textcolor{grey}{52.9} \\
    \textcolor{grey}{Qwen2-VL-7B~\cite{wang2024qwen2vlenhancingvisionlanguagemodels}} & \textcolor{grey}{64} & \textcolor{grey}{69.1} & \textcolor{grey}{53.2} & \textcolor{grey}{63.8} & \textcolor{grey}{50.6} & \textcolor{grey}{66.3} & \textcolor{grey}{60.9} & \textcolor{grey}{60.7} & \textcolor{grey}{44.4} & \textcolor{grey}{66.9} & \textcolor{grey}{34.4} & \textcolor{grey}{48.6} & \textcolor{grey}{30.1} & \textcolor{grey}{65.7} & \textcolor{grey}{50.8} & \textcolor{grey}{48.9} & \textcolor{grey}{52.7} \\
    \textcolor{grey}{InternVL-V2-8B~\cite{chen2024internvl}} & \textcolor{grey}{64} & \textcolor{grey}{68.5} & \textcolor{grey}{58.7} & \textcolor{grey}{69.0} & \textcolor{grey}{44.9} & \textcolor{grey}{67.3} & \textcolor{grey}{56.0} & \textcolor{grey}{60.7} & \textcolor{grey}{43.1} & \textcolor{grey}{61.5} & \textcolor{grey}{27.4} & \textcolor{grey}{44.0} & \textcolor{grey}{25.8} & \textcolor{grey}{57.6} & \textcolor{grey}{52.9} & \textcolor{grey}{45.4} & \textcolor{grey}{50.1} \\
    \textcolor{grey}{LongVU-7B~\cite{shen2024longvu}} & \textcolor{grey}{1fps} & \textcolor{grey}{55.7} & \textcolor{grey}{49.5} & \textcolor{grey}{59.5} & \textcolor{grey}{48.3} & \textcolor{grey}{68.3} & \textcolor{grey}{63.0} & \textcolor{grey}{57.4} & \textcolor{grey}{43.1} & \textcolor{grey}{66.2} & \textcolor{grey}{9.1} & \textcolor{grey}{39.5} & \textcolor{grey}{16.6} & \textcolor{grey}{69.0} & \textcolor{grey}{60.0} & \textcolor{grey}{48.5} & \textcolor{grey}{48.5} \\
    \midrule
    \multicolumn{18}{c}{\textbf{Open-source Online Video-LLMs}} \\
    \midrule
    Flash-VStream-7B~\cite{zhang2024flash}~\textsubscript{\textcolor{grey}{[ArXiv 2024]}} & 1fps & 25.5 & 32.1 & 29.3 & 33.7 & 29.7 & 28.8 & 29.9 & 36.4 & 33.8 & 5.9 & 25.4 & 5.4 & \uline{67.3} & \uline{60.0} & \uline{44.2} & 33.2 \\
    VideoLLM-online-8B~\cite{chen2024videollm}~\textsubscript{\textcolor{grey}{[CVPR 2024]}} & 2fps & 8.1 & 23.9 & 12.1 & 14.0 & 45.5 & 21.2 & 20.8 & 22.2 & 18.8 & \textbf{12.2} & 17.7 & - & - & - & - & - \\
    TimeChat-Online-7B~\cite{yao2025timechat}~\textsubscript{\textcolor{grey}{[MM 2025]}} & 1fps & \uline{75.2} & \uline{46.8} & \uline{70.7} & \uline{47.8} & \uline{69.3} & \uline{61.4} & \uline{61.9} & \uline{55.9} & \uline{59.5} & 9.7 & \uline{41.7} & \uline{31.6} & 38.5 & 40.0 & 36.7 & \uline{46.7} \\
    \midrule
    \rowcolor{blue!8}
    \rowcolor{blue!8}
    \textbf{WAT} & 1fps & \textbf{81.2} & \textbf{48.3} & \textbf{70.7} & \textbf{48.3} & \textbf{74.0} & \textbf{65.2} & \textbf{64.7} & \textbf{58.9} & \textbf{64.9} & \uline{11.8} & \textbf{45.2} & \textbf{36.9} & \textbf{70.1} & \textbf{60.4} & \textbf{55.8} & \textbf{55.2} \\
    \bottomrule
    \end{tabular}
}
\end{table*}

%% file: table/offline.tex
\begin{table}[t]
    \centering
    \caption{\small Results on offline long video benchmarks. We report the accuracy on the MLVU multiple-choice questions, and VideoMME without subtitles. The \textbf{bold} values indicate the best performance, and \uline{underlined} values indicate the second best.}
    \label{tab:long_video_benchmarks}
\adjustbox{width=\linewidth}
  {
  \setlength{\tabcolsep}{0.5pt}
    \begin{tabular}{lc|ccc}
    \toprule
    \multirow{2}{*}{\textbf{Model}} & \multirow{2}{*}{\textbf{\# Frames}} & \multirow{2}{*}{\textbf{MLVU}} & \multicolumn{2}{c}{\textbf{VideoMME}} \\
    \cmidrule(lr){4-5}
    \addlinespace[1pt]
    & & & overall & long \\
    \midrule
    \multirow{2}{*}{\textbf{Video Length}} & \multirow{2}{*}{-} & \makecell{3--120} & \makecell{1--60} & \makecell{30--60} \\
    & & min & min & min \\
    \midrule
    \multicolumn{5}{c}{\textbf{Open-Source Offline VideoLLMs}} \\
    \midrule
    LLaMA-VID-7B~\cite{li2024llama} & 1 fps & 33.2 & - & - \\
    MovieChat-7B~\cite{song2024moviechat} & 2048 & 25.8 & 38.2 & 33.4 \\
    LLaVA-Next-Video-7B~\cite{zhang2024video} & 32 & - & 46.6 & - \\
    VideoChat2-7B~\cite{li2023mvbench} & 16 & 47.9 & 39.5 & 33.2 \\
    LongVA-7B~\cite{zhang2024longva} & 128 & 56.3 & 52.6 & 46.2 \\
    Kangaroo-7B~\cite{kangaroogroup} & 64 & 61.0 & 56.0 & 46.6 \\
    Video-CCAM-14B~\cite{fei2024video} & 96 & 63.1 & 53.2 & 46.7 \\
    VideoXL-7B~\cite{shu2025video} & 128 & 64.9 & 55.5 & 49.2 \\
    Qwen2.5-VL-7B~\cite{xu2025qwen2} & 1 fps & 66.9 & 63.2 & 50.4 \\
    \midrule
    \multicolumn{5}{c}{\textbf{Open-source Online  VideoLLMs}} \\
    \midrule
    Dispider-7B~\cite{qian2025dispider}~\textsubscript{\textcolor{grey}{[CVPR 2025]}} & 1 fps & 61.7 & 57.2 & - \\
    VideoChat-Online-8B~\cite{chen2024videollm}~\textsubscript{\textcolor{grey}{[CVPR 2025]}} & 2 fps & - & 52.8 & 44.9 \\ 
    TimeChat-Online-7B~\cite{yao2025timechat}~\textsubscript{\textcolor{grey}{[MM 2025]}} & 1 fps & \textbf{65.4} & \textbf{62.5} & \uline{49.2} \\
    \midrule
    \rowcolor{blue!8}\textbf{WAT} & 1 fps & \uline{63.9} & \uline{62.4} & \textbf{50.8} \\
    \bottomrule
    \end{tabular}
}
\end{table}

%% file: table/ab_memory.tex
\begin{table}[t]
    \centering
    \caption{Ablation study of hierarchical memory system.}
    \label{tab:memory}
\adjustbox{width=\linewidth}
  {
  \setlength{\tabcolsep}{12pt}
    \begin{tabular}{lcc}
    \toprule
    \textbf{Model} & \textbf{OVO-Bench} & \textbf{StreamingBench} \\
    \midrule
    Qwen2.5-VL & - & 73.7 \\
    WAT (Only LTM) & 50.3 & 75.7 \\
    WAT (STM+LTM)  & \textbf{55.2} & \textbf{77.7} \\
    \bottomrule
    \end{tabular}
    }
\end{table}

%% file: table/ab_ltm.tex
\begin{table}[t]
    \centering
    \caption{Performance comparison with different LTM lengths. Relative gain is compared with $N_{\mathrm{LTM}}=32$.}
    \label{tab:ltm}
\adjustbox{width=\linewidth}
  {
  \setlength{\tabcolsep}{7pt}
    \resizebox{\columnwidth}{!}{
    \begin{tabular}{lcccc}
        \toprule
        \textbf{$N_{\mathrm{LTM}}$} &
        \textbf{OVO-Bench} &
        Gain &
        \textbf{StreamingBench} &
        Gain \\
        \midrule
        32  & 52.0 & --   & 76.9 & --   \\
        128 & 52.6 & +0.6 & 77.1 & +0.2 \\
        256 & 53.7 & +1.7 & 77.2 & +0.3 \\
        768 & \textbf{55.2} & \textbf{+3.2} & \textbf{77.7} & \textbf{+0.8} \\
        \bottomrule
    \end{tabular}}
}
\end{table}

%% file: table/ab_racl.tex
\begin{table}[t]
    \centering
    \caption{Ablation study on the loss composition. We denote the next token prediction loss, the retrieval–alignment contrastive learning loss, and the softmax temperature parameter as $\mathcal{L}_{\text{NTP}}$, $\mathcal{L}_{\text{RACL}}$, and $\tau$, respectively.}
    \label{tab:racl}
\adjustbox{width=\linewidth}
  {
  \setlength{\tabcolsep}{7pt}
    \begin{tabular}{lccc}
    \toprule
    \textbf{Loss}&\textbf{$\tau$}&\textbf{OVO-Bench} &\textbf{StreamingBench} \\
    \midrule
    $\mathcal{L}_{\text{NTP}}$ &- &50.1 &76.4 \\
    $\mathcal{L}_{\text{NTP}}$ + $\mathcal{L}_{\text{RACL}}$ & 0.01 &54.6 &77.3 \\
    $\mathcal{L}_{\text{NTP}}$ + $\mathcal{L}_{\text{RACL}}$ & 0.07 &55.2 &77.7 \\
    \bottomrule
    \end{tabular}
}
\end{table}

%% file: bare_jrnl_new_sample4.bbl
\begin{thebibliography}{1}
\bibliographystyle{IEEEtran}

\bibitem{an2026aiflow}
H.~An, W.~Hu, S.~Huang, \textit{et al.},
``AI flow: Perspectives, scenarios, and approaches,''
\textit{Vicinagearth}, vol.~3, no.~1, 2026, doi:~10.1007/s44336-025-00031-y.

\bibitem{li2024multiagent}
X.~Li, S.~Wang, S.~Zeng, \textit{et al.},
``A survey on LLM-based multi-agent systems: Workflow, infrastructure, and challenges,''
\textit{Vicinagearth}, vol.~1, no.~9, 2024, doi:~10.1007/s44336-024-00009-2.

\bibitem{shen2025lvgvos}
Y.~Shen and D.~Zhang,
``A survey of language-guided video object segmentation: From referring to reasoning,''
\textit{Vicinagearth}, vol.~2, no.~9, 2025, doi:~10.1007/s44336-025-00018-9.


\bibitem{liu2025trainingfree}
J.~Liu, Y.~Wang, L.~Zhang, \textit{et al.},
``Towards training-free long video understanding: Methods, benchmarks, and open challenges,''
\textit{Vicinagearth}, vol.~2, no.~6, 2025, doi:~10.1007/s44336-025-00017-w.

\bibitem{yu2024multi}
T.~Yu, K.~Fu, J.~Zhang, Q.~Huang, and J.~Yu,
``Multi-granularity contrastive cross-modal collaborative generation for end-to-end long-term video question answering,''
\textit{IEEE Trans. Image Process.}, vol.~33, pp.~3115--3129, 2024.

\bibitem{fu2025boosting}
Z.~Fu, Z.~Mao, L.~Zhang, and Y.~Zhang,
``Boosting faithful multimodal LLMs via complementary visual grounding,''
\textit{IEEE Trans. Image Process.}, vol.~34, pp.~8641--8655, 2025.

\bibitem{xie2025caption}
P.~Xie, J.~Li, G.~Lu, Y.~Xu, and D.~Zhang,
``Caption assisted multimodal large language model for video moment retrieval,''
\textit{IEEE Trans. Image Process.}, 2025.

\bibitem{yellinek20253vl}
N.~Yellinek, L.~Karlinsky, and R.~Giryes,
``3VL: Using trees to improve vision-language models' interpretability,''
\textit{IEEE Trans. Image Process.}, 2025.



\bibitem{ataallah2024minigpt4}
K.~Ataallah, X.~Shen, E.~Abdelrahman, E.~Sleiman, D.~Zhu, J.~Ding, and M.~Elhoseiny,
``Minigpt4-video: Advancing multimodal LLMs for video understanding with interleaved visual-textual tokens,''
\textit{arXiv preprint arXiv:2404.03413}, 2024.

\bibitem{bai2025qwen25vltechnicalreport}
S.~Bai \emph{et al.},
``Qwen2.5-VL technical report,''
\textit{arXiv preprint arXiv:2502.13923}, 2025.

\bibitem{li2025videochat}
K.~Li, Y.~He, Y.~Wang, Y.~Li, W.~Wang, P.~Luo, Y.~Wang, L.~Wang, and Y.~Qiao,
``Videochat: Chat-centric video understanding,''
\textit{Sci. China Inf. Sci.}, vol.~68, no.~10, p.~200102, 2025.

\bibitem{li2024temporal}
L.~Li, Y.~Liu, L.~Yao, P.~Zhang, C.~An, L.~Wang, X.~Sun, L.~Kong, and Q.~Liu,
``Temporal reasoning transfer from text to video,''
\textit{arXiv preprint arXiv:2410.06166}, 2024.

\bibitem{liu2024llavanext}
H.~Liu \emph{et al.},
``Llavanext: Improved reasoning, OCR, and world knowledge,''
\textit{arXiv preprint}, 2024.

\bibitem{ren2024timechat}
S.~Ren, L.~Yao, S.~Li, X.~Sun, and L.~Hou,
``Timechat: A time-sensitive multimodal large language model for long video understanding,''
in \textit{Proc. IEEE/CVF Conf. Comput. Vis. Pattern Recognit. (CVPR)},
pp.~14313--14323, 2024.

\bibitem{zhang2025videollama}
B.~Zhang \emph{et al.},
``Videollama 3: Frontier multimodal foundation models for image and video understanding,''
\textit{arXiv preprint arXiv:2501.13106}, 2025.

\bibitem{zhang2024internlm}
P.~Zhang \emph{et al.},
``InternLM-XComposer2.5-OmniLive: A comprehensive multimodal system for long-term streaming video and audio interactions,''
\textit{arXiv preprint arXiv:2412.09596}, 2024.

\bibitem{zhang2024video}
Y.~Zhang, J.~Wu, W.~Li, B.~Li, Z.~Ma, Z.~Liu, and C.~Li,
``Video instruction tuning with synthetic data,''
\textit{arXiv preprint arXiv:2410.02713}, 2024.

\bibitem{xu2024slowfast}
M.~Xu, M.~Gao, Z.~Gan, H.-Y.~Chen, Z.~Lai, H.~Gang, K.~Kang, and A.~Dehghan,
``SlowFast-LLaVA: A strong training-free baseline for video large language models,''
\textit{arXiv preprint arXiv:2407.15841}, 2024.

\bibitem{shao2024lmdrive}
H.~Shao, Y.~Hu, L.~Wang, G.~Song, S.~L.~Waslander, Y.~Liu, and H.~Li,
``LMDrive: Closed-loop end-to-end driving with large language models,''
in \textit{Proc. IEEE/CVF Conf. Comput. Vis. Pattern Recognit. (CVPR)},
pp.~15120--15130, 2024.

\bibitem{chen2025livecc}
J.~Chen, Z.~Zeng, Y.~Lin, W.~Li, Z.~Ma, and M.~Z.~Shou,
``LiveCC: Learning video LLMs with streaming speech transcription at scale,''
in \textit{Proc. IEEE/CVF Conf. Comput. Vis. Pattern Recognit. (CVPR)},
pp.~29083--29095, 2025.

\bibitem{zhu2024spa}
H.~Zhu, H.~Yang, Y.~Wang, J.~Yang, L.~Wang, and T.~He,
``SPA: 3D spatial-awareness enables effective embodied representation,''
\textit{arXiv preprint arXiv:2410.08208}, 2024.


\bibitem{chen2024videollm}
J.~Chen, Z.~Lv, S.~Wu, K.~Q.~Lin, C.~Song, D.~Gao, J.-W.~Liu, Z.~Gao, D.~Mao, and M.~Z.~Shou,
``Videollm-online: Online video large language model for streaming video,''
in \textit{Proc. IEEE/CVF Conf. Comput. Vis. Pattern Recognit. (CVPR)},
pp.~18407--18418, 2024.

\bibitem{di2025streaming}
S.~Di, Z.~Yu, G.~Zhang, H.~Li, T.~Zhong, H.~Cheng, B.~Li, W.~He, F.~Shu, and H.~Jiang,
``Streaming video question-answering with in-context video KV-cache retrieval,''
\textit{arXiv preprint arXiv:2503.00540}, 2025.

\bibitem{qian2024streaming}
R.~Qian, X.~Dong, P.~Zhang, Y.~Zang, S.~Ding, D.~Lin, and J.~Wang,
``Streaming long video understanding with large language models,''
\textit{Adv. Neural Inf. Process. Syst.}, vol.~37, pp.~119336--119360, 2024.

\bibitem{xu2025qwen2}
J.~Xu \emph{et al.},
``Qwen2.5-Omni technical report,''
\textit{arXiv preprint arXiv:2503.20215}, 2025.

\bibitem{zhang2024flash}
H.~Zhang, Y.~Wang, Y.~Tang, Y.~Liu, J.~Feng, J.~Dai, and X.~Jin,
``Flash-vstream: Memory-based real-time understanding for long video streams,''
\textit{arXiv preprint arXiv:2406.08085}, 2024.

\bibitem{zhou2024mlvu}
J.~Zhou, Y.~Shu, B.~Zhao, B.~Wu, S.~Xiao, X.~Yang, Y.~Xiong, B.~Zhang,
T.~Huang, and Z.~Liu,
``MLVU: A comprehensive benchmark for multi-task long video understanding,''
\textit{arXiv preprint arXiv:2406.04264}, 2024.

\bibitem{fu2025video}
C.~Fu, Y.~Dai, Y.~Luo, L.~Li, S.~Ren, R.~Zhang, Z.~Wang, C.~Zhou,
Y.~Shen, M.~Zhang, \textit{et al.},
``Video-MME: The first-ever comprehensive evaluation benchmark of multimodal LLMs in video analysis,''
in \textit{Proc. IEEE/CVF Conf. Comput. Vis. Pattern Recognit. (CVPR)},
pp.~24108--24118, 2025.


\bibitem{qian2025dispider}
R.~Qian, S.~Ding, X.~Dong, P.~Zhang, Y.~Zang, Y.~Cao, D.~Lin, and J.~Wang,
``Dispider: Enabling video LLMs with active real-time interaction via disentangled perception, decision, and reaction,''
in \textit{Proc. IEEE/CVF Conf. Comput. Vis. Pattern Recognit. (CVPR)},
pp.~24045--24055, 2025.

\bibitem{yao2025timechat}
L.~Yao \emph{et al.},
``Timechat-online: 80\% visual tokens are naturally redundant in streaming videos,''
in \textit{Proc. 33rd ACM Int. Conf. Multimedia (ACM MM)},
pp.~10807--10816, 2025.

\bibitem{cheng2024videollama}
Z.~Cheng, S.~Leng, H.~Zhang, Y.~Xin, X.~Li, G.~Chen, Y.~Zhu, W.~Zhang, Z.~Luo, D.~Zhao, \textit{et al.},
``VideoLLaMA 2: Advancing spatial-temporal modeling and audio understanding in video-LLMs,''
\textit{arXiv preprint arXiv:2406.07476}, 2024.

\bibitem{tang2025adaptive}
X.~Tang, J.~Qiu, L.~Xie, Y.~Tian, J.~Jiao, and Q.~Ye,
``Adaptive keyframe sampling for long video understanding,''
in \textit{Proc. IEEE/CVF Conf. Comput. Vis. Pattern Recognit. (CVPR)},
pp.~29118--29128, 2025.

\bibitem{radford2021learning}
A.~Radford \emph{et al.},
``Learning transferable visual models from natural language supervision,''
in \textit{Proc. Int. Conf. Mach. Learn. (ICML)},
pp.~8748--8763, 2021.

\bibitem{tang2025tspo}
C.~Tang, Z.~Han, H.~Sun, S.~Zhou, X.~Zhang, X.~Wei, Y.~Yuan, H.~Zhang, J.~Xu, and H.~Sun,
``TSPO: Temporal sampling policy optimization for long-form video language understanding,''
\textit{arXiv preprint arXiv:2508.04369}, 2025.

\bibitem{hu2025cos}
J.~Hu, Z.~Cheng, C.~Si, W.~Li, and S.~Gong,
``CoS: Chain-of-shot prompting for long video understanding,''
\textit{arXiv preprint arXiv:2502.06428}, 2025.

\bibitem{liu2024streamchat}
J.~Liu, Z.~Yu, S.~Lan, S.~Wang, R.~Fang, J.~Kautz, H.~Li, and J.~M.~Alvare,
``StreamChat: Chatting with streaming video,''
\textit{arXiv preprint arXiv:2412.08646}, 2024.

\bibitem{huang2025online}
Z.~Huang, X.~Li, J.~Li, J.~Wang, X.~Zeng, C.~Liang, T.~Wu, X.~Chen, L.~Li, and L.~Wang,
``Online video understanding: OVBench and VideoChat-Online,''
in \textit{Proc. IEEE/CVF Conf. Comput. Vis. Pattern Recognit. (CVPR)},
pp.~3328--3338, 2025.

\bibitem{di2025streaming}
S.~Di, Z.~Yu, G.~Zhang, H.~Li, T.~Zhong, H.~Cheng, B.~Li, W.~He, F.~Shu, and H.~Jiang,
``Streaming video question-answering with in-context video KV-cache retrieval,''
\textit{arXiv preprint arXiv:2503.00540}, 2025.

\bibitem{ning2024inf}
Z.~Ning, J.~Zhao, Q.~Jin, W.~Ding, and M.~Guo,
``Inf-MLLM: Efficient streaming inference of multimodal large language models on a single GPU,''
\textit{arXiv preprint arXiv:2409.09086}, 2024.

\bibitem{wu2024videollm}
S.~Wu, J.~Chen, K.~Q.~Lin, Q.~Wang, Y.~Gao, Q.~Xu, T.~Xu, Y.~Hu, E.~Chen, and M.~Z.~Shou,
``VideoLLM-Mod: Efficient video-language streaming with mixture-of-depths vision computation,''
\textit{Adv. Neural Inf. Process. Syst.}, vol.~37, pp.~109922--109947, 2024.

\bibitem{wang2024videollm}
Y.~Wang, X.~Meng, Y.~Wang, J.~Liang, J.~Wei, H.~Zhang, and D.~Zhao,
``VideoLLM knows when to speak: Enhancing time-sensitive video comprehension with video-text duet interaction format,''
\textit{arXiv preprint arXiv:2411.17991}, 2024.

\bibitem{lin2024streamingbench}
J.~Lin, Z.~Fang, C.~Chen, Z.~Wan, F.~Luo, P.~Li, Y.~Liu, and M.~Sun,
``StreamingBench: Assessing the gap for MLLMs to achieve streaming video understanding,''
\textit{arXiv preprint arXiv:2411.03628}, 2024.


\bibitem{niu2025ovo}
J.~Niu, Y.~Li, Z.~Miao, C.~Ge, Y.~Zhou, Q.~He, X.~Dong, H.~Duan, S.~Ding, R.~Qian, \emph{et~al.},
``OVO-Bench: How far is your video-LLMs from real-world online video understanding?''
in \textit{Proc. IEEE/CVF Conf. Comput. Vis. Pattern Recognit. (CVPR)},
pp.~18902--18913, 2025.

\bibitem{team2024gemini}
Gemini~Team, P.~Georgiev, V.~I.~Lei, R.~Burnell, L.~Bai, A.~Gulati,
G.~Tanzer, D.~Vincent, Z.~Pan, S.~Wang, \textit{et al.},
``Gemini~1.5: Unlocking multimodal understanding across millions of tokens of context,''
\textit{arXiv preprint arXiv:2403.05530}, 2024.

\bibitem{hurst2024gpt}
A.~Hurst, A.~Lerer, A.~P.~Goucher, A.~Perelman, A.~Ramesh, A.~Clark,
A.~J.~Ostrow, A.~Welihinda, A.~Hayes, A.~Radford, \textit{et al.},
``GPT-4o system card,''
\textit{arXiv preprint arXiv:2410.21276}, 2024.

\bibitem{anthropic2024claude35sonnet}
Anthropic,
``Claude 3.5 Sonnet,''
2024. [Online]. Available: https://www.anthropic.com/claude

\bibitem{cheng2024videollama}
Z.~Cheng, S.~Leng, H.~Zhang, Y.~Xin, X.~Li, G.~Chen, Y.~Zhu, W.~Zhang, Z.~Luo, D.~Zhao, \textit{et al.},
``VideoLLaMA 2: Advancing spatial-temporal modeling and audio understanding in video-LLMs,''
\textit{arXiv preprint arXiv:2406.07476}, 2024.

\bibitem{lin2023vila}
J.~Lin, H.~Yin, W.~Ping, Y.~Lu, P.~Molchanov, A.~Tao, H.~Mao, J.~Kautz, M.~Shoeybi, and S.~Han,
``VILA: On pre-training for visual language models,''
\textit{arXiv preprint arXiv:2312.07533}, 2023.

\bibitem{fei2024video}
J.~Fei, D.~Li, Z.~Deng, Z.~Wang, G.~Liu, and H.~Wang,
``Video-CCAM: Enhancing video-language understanding with causal cross-attention masks for short and long videos,''
\textit{arXiv preprint arXiv:2408.14023}, 2024.

\bibitem{zhang2024longva}
P.~Zhang, K.~Zhang, B.~Li, G.~Zeng, J.~Yang, Y.~Zhang, Z.~Wang, H.~Tan,
C.~Li, and Z.~Liu,
``Long context transfer from language to vision,''
\textit{arXiv preprint arXiv:2406.16852}, 2024.

\bibitem{chen2024internvl}
Z.~Chen, J.~Wu, W.~Wang, W.~Su, G.~Chen, S.~Xing, M.~Zhong, Q.~Zhang,
X.~Zhu, L.~Lu, \textit{et al.},
``InternVL: Scaling up vision foundation models and aligning for generic visual-linguistic tasks,''
in \textit{Proc. IEEE/CVF Conf. Comput. Vis. Pattern Recognit. (CVPR)},
pp.~24185--24198, 2024.

\bibitem{kangaroogroup}
J.~Liu, Y.~Wang, H.~Ma, X.~Wu, X.~Ma, X.~Wei, J.~Jiao, E.~Wu, and J.~Hu,
``Kangaroo: A powerful video-language model supporting long-context video input,''
\textit{arXiv preprint arXiv:2408.15542}, 2024.

\bibitem{yao2024minicpm}
Y.~Yao, T.~Yu, A.~Zhang, C.~Wang, J.~Cui, H.~Zhu, T.~Cai, H.~Li,
W.~Zhao, Z.~He, \textit{et al.},
``MiniCPM-V: A GPT-4V level MLLM on your phone,''
\textit{arXiv preprint arXiv:2408.01800}, 2024.


\bibitem{li2024llava}
B.~Li, Y.~Zhang, D.~Guo, R.~Zhang, F.~Li, H.~Zhang, K.~Zhang, P.~Zhang,
Y.~Li, Z.~Liu, \textit{et al.},
``LLaVA-OneVision: Easy visual task transfer,''
\textit{arXiv preprint arXiv:2408.03326}, 2024.

\bibitem{wang2024qwen2vlenhancingvisionlanguagemodels}
P.~Wang, S.~Bai, S.~Tan, S.~Wang, Z.~Fan, J.~Bai, K.~Chen, X.~Liu,
J.~Wang, W.~Ge, Y.~Fan, K.~Dang, M.~Du, X.~Ren, R.~Men, D.~Liu,
C.~Zhou, J.~Zhou, and J.~Lin,
``Qwen2-VL: Enhancing vision-language models' perception of the world at any resolution,''
\textit{arXiv preprint arXiv:2409.12191}, 2024.

\bibitem{shen2024longvu}
X.~Shen, Y.~Xiong, C.~Zhao, L.~Wu, J.~Chen, C.~Zhu, Z.~Liu, F.~Xiao,
B.~Varadarajan, F.~Bordes, \textit{et al.},
``LongVU: Spatiotemporal adaptive compression for long video-language understanding,''
\textit{arXiv preprint arXiv:2410.17434}, 2024.

\bibitem{li2024llama}
Y.~Li, C.~Wang, and J.~Jia,
``LLaMA-VID: An image is worth 2 tokens in large language models,''
in \textit{Proc. Eur. Conf. Comput. Vis. (ECCV)},
pp.~323--340, 2024.

\bibitem{song2024moviechat}
E.~Song, W.~Chai, G.~Wang, Y.~Zhang, H.~Zhou, F.~Wu, H.~Chi, X.~Guo,
T.~Ye, Y.~Zhang, \textit{et al.},
``MovieChat: From dense token to sparse memory for long video understanding,''
in \textit{Proc. IEEE/CVF Conf. Comput. Vis. Pattern Recognit. (CVPR)},
pp.~18221--18232, 2024.

\bibitem{li2023mvbench}
K.~Li, Y.~Wang, Y.~He, Y.~Li, Y.~Wang, Y.~Liu, Z.~Wang, J.~Xu, G.~Chen,
P.~Luo, L.~Wang, and Y.~Qiao,
``MVBench: A comprehensive multi-modal video understanding benchmark,''
\textit{arXiv preprint arXiv:2311.17005}, 2023.

\bibitem{shu2025video}
Y.~Shu, Z.~Liu, P.~Zhang, M.~Qin, J.~Zhou, Z.~Liang, T.~Huang, and B.~Zhao,
``Video-XL: Extra-long vision language model for hour-scale video understanding,''
in \textit{Proc. IEEE/CVF Conf. Comput. Vis. Pattern Recognit. (CVPR)},
pp.~26160--26169, 2025.


\end{thebibliography}
